\documentclass[preprint,authoryear]{elsarticle}

\usepackage{graphicx}
\usepackage{algorithm}
\usepackage{algpseudocode}
\usepackage{amsmath}
\usepackage{amssymb}
\usepackage{bm}
\usepackage{color} 

\journal{Neural Networks}

\newcommand{\bx}{\bm{x}}

\newcommand{\bu}{\bm{u}}
\newcommand{\bw}{\bm{w}}

\begin{document}

\begin{frontmatter}

\title{Forward and inverse reinforcement learning sharing network weights and hyperparameters}

\author[ATR]{Eiji Uchibe\corref{cor1}}
\ead{uchibe@atr.jp}
\cortext[cor1]{Corresponding author.}

\author[OIST]{Kenji Doya}
\ead{doya@oist.jp}

\address[ATR]{Department of Brain Robot Interface,
 ATR Computational Neuroscience Laboratories,
 2-2-2 Hikaridai, Seika-cho, Soraku-gun, Kyoto 619-0288, Japan}
\address[OIST]{Neural Computation Unit, Okinawa Institute of 
 Science and Technology Graduate University,
 1919-1 Tancha, Onna-son, Okinawa 904-0495, Japan}

\begin{abstract}
  This paper proposes model-free imitation learning named
  Entropy-Regularized Imitation Learning (ERIL)
  that minimizes the reverse Kullback-Leibler (KL)
    divergence. ERIL
  combines forward and inverse reinforcement learning
  (RL) under the
  framework of an entropy-regularized Markov decision process.
  An inverse RL step computes
  the log-ratio between two distributions by
  evaluating
  two binary discriminators.
  The first discriminator distinguishes the state generated by
  the forward RL step from the expert's state.
  The second discriminator, which is structured by the theory of
    entropy regularization, distinguishes the
    state-action-next-state tuples generated by the learner from
    the expert ones.
  One notable feature is that
  the second discriminator shares hyperparameters with the
  forward RL, which can be used to
  control the discriminator's ability.
  A forward RL step
  minimizes the reverse KL estimated by
  the inverse RL step.
  We show that minimizing the reverse KL divergence is equivalent
  to finding an optimal policy.
  Our experimental results on MuJoCo-simulated environments
  and vision-based reaching tasks with a robotic arm
  show that
  ERIL is more sample-efficient than the baseline methods.
  We apply the method to human behaviors that perform a
  pole-balancing task and describe how the estimated reward functions
  show how every subject achieves her
  goal.\par
\end{abstract}


\end{frontmatter}


\section{Introduction}
\label{sec:introduction}

Reinforcement Learning (RL) is a computational framework for
investigating the decision-making processes of both biological and
artificial systems that can learn an optimal policy by interacting
with an environment \citep{Sutton1998a,Doya2007a,Kober2013a}.
  Modern RL algorithms have achieved remarkable performance in
  playing Atari games \citep{Mnih2015a}, Go \citep{Silver2017a},
  Dota 2 \citep{OpenAI2019a}, and Starcraft II \citep{Vinyals2019a}.
They have also been successfully applied to dexterous manipulation
  tasks \citep{OpenAI2019b}, folding a T-shirt \citep{Tsurumine2019a},
  quadruped locomotion \citep{Haarnoja2018c}, the optimal state
  feedback control of nonaffine nonlinear systems \citep{Wang2019a},
  and non-zero-sum game output regulation problems \citep{Odekunle2020a}. 
However, one critical open question in RL is designing and
preparing an appropriate reward function for a given task. 
Although it is easy to design a sparse reward function that gives a 
positive reward when a task is accomplished and zero otherwise,
such an approach complicates finding an optimal policy due to 
prohibitive learning times.
On the other hand, we can accelerate the learning speed with a
complicated function that generally gives a non-zero reward signal.
However, optimized behaviors often deviate from an
experimenter's intention if the reward function is too complicated
\citep{Doya2005a}. 

In some situations, it is easier to prepare examples of a desired behavior
provided by an expert
than handcrafting an appropriate reward function.
  Behavior Cloning (BC) is a straightforward approach in imitation
  learning formulated as supervised learning. BC
minimizes the forward Kullback-Leibler
(KL)
divergence, which is known as moment projection (M-projection).
Forward KL is the expectation of the log-likelihood ratio
under expert distribution. Although BC requires no
interaction with the environment, it suffers from a state covariate
shift
  problem: small errors in actions introduce the learner to unseen
  states that are not included in the training dataset \citep{Ross2011a}.
  Also, BC shows poor performance
if the model is misspecified
because minimizing the forward KL has a mode-covering property.
To overcome the covariate shift problem,
several inverse RL
\citep{Ng2000b} and apprenticeship learning \citep{Abbeel2004a} methods have
been proposed to retrieve a reward function from 
expert behaviors and implement imitation learning. 
Currently available applications include
a probabilistic driver route prediction system 
\citep{Vogel2012a,Liu2013b},
modeling risk anticipation and defensive driving 
\citep{Shimosaka2014a}, investigating human behaviors in table tennis
\citep{Muelling2014a}, robot navigation tasks 
\citep{Kretzschmar2016a,Xia2016a}, analyzing animal behaviors
\citep{Ashida2019a,Hirakawa2018a,Yamaguchi2018a}, 
and parser training \citep{Neu2009a}.
A recent functional magnetic resonance imaging (fMRI) study suggests
that the anterior part of the dorsomedial prefrontal cortex (dmPFC) is
likely to encode the inverse reinforcement learning algorithm
\citep{Collette2017a}.
Combining inverse RL with a standard RL is a promising approach to
find an optimal policy from expert demonstrations. Hereafter,
we use the term ``forward'' reinforcement learning to clarify the
difference.

Recently, some works \citep{Fu2018a,Ho2016c} have connected forward
and inverse RL and Generative Adversarial Networks (GANs)
\citep{Goodfellow2014a}, 
which exhibited remarkable success in image 
generation, video prediction, and machine translation domains. 
In this view, inverse RL is interpreted as a GAN discriminator whose
goal is to determine whether experiences are drawn from an expert or 
generated by a forward RL step.
The GAN generator is implemented by a forward RL step and creates experiences that are indistinguishable by an inverse RL. 
Generative Adversarial Imitation Learning (GAIL) \citep{Ho2016c} 
showed that the iterative process of forward and inverse RL produces
policies that outperformed BC. 
However, GAIL minimizes the Jensen-Shannon divergence, which has a similar
forward KL property.
  In addition, GAIL is sample-inefficient because an on-policy
  RL algorithm used in the forward RL step simply trains a policy
  from a reward calculated by the inverse RL step.
  To improve the sample efficiency in the forward RL step,
  \citet{Jena2020a} added BC loss to the loss of the GAIL generator.
  \citet{Kinose2020a} integrated the GAIL discriminator with
  reinforcement learning, in which the policy is trained with
  both the original reward and additional rewards calculated
  by the discriminator.
  However, their approaches remain sample-inefficient.
  Utilizing the result of the inverse RL step to the forward
  RL step and vice versa is difficult because the discriminator's structure is designed independently from the
  generator.

To further improve the sample efficiency,
this paper proposes
a model-free imitation learning algorithm named
Entropy-Regularized Imitation Learning (ERIL), which minimizes the
information projection or the I-projection, which is also known as
the reverse KL divergence between two probability
distributions induced by a learner and an expert.
A reverse KL, which is the expectation of the log-likelihood
ratio under the learner's distribution, has a mode-seeking
property that focuses on the distribution mode that the
policy can represent. A reverse KL is more appropriate than a forward KL
when the policy is misspecified.
Unfortunately, reverse KL divergence cannot be computed
because the expert distribution is unknown.
Our idea applies the density ratio trick \citep{Sugiyama2012b} to
evaluate the log-ratio between two distributions from samples drawn
from them.
In addition, we exploit the framework of the entropy-regularized Markov
Decision Process (MDP), where the reward function is augmented by the
differential entropy of a learner's policy and the KL divergence between
the learner and expert policies.
Consequently, the log-ratio can be computed by training two
binary discriminators.
One is a state-only discriminator, which distinguishes the state
generated by the learner from the expert's state.
The second discriminator, which is a function of a tuple of a state, an action, and the
next state, also distinguishes between the experiences of learners
and experts.
The second discriminator is represented by reward, a state value function,
and the log-ratio of the first discriminator.
We show that Adversarial Inverse Reinforcement Learning (AIRL)
\citep{Fu2018a} and
Logistic Regression-based Inverse RL (LogReg-IRL)
\citep{Uchibe2014b,Uchibe2018c} discriminators
are a special case of an ERIL.
The loss function is essentially identical as that of GAN for training
discriminators, which are efficiently trained by logistic regression.

After evaluating the log-ratio, the forward RL in ERIL minimizes the
estimated I-projection.
We show that its minimization is equivalent to maximizing the
entropy-regularized reward. Consequently, the forward RL algorithm is
implemented by off-policy reinforcement learning that resembles
Dynamic Policy Programming \citep{Azar2012a}, Soft Actor-Critic
(SAC)
\citep{Haarnoja2018a}, and conservative value iteration \citep{Kozuno2019a}.
  In the forward RL step, the state value, the state-action value,
  and the stochastic policy are trained by the actor-critic algorithm,
 and the reward function estimated by the inverse RL step is
  fixed. This step allows the learner to generalize the expert
  policy to unseen states that are not included in the demonstrations.

We experimented with the MuJoCo benchmark tasks 
\citep{Todorov2012a} in the OpenAI gym \citep{Brockman2016a}.
Our experimental results demonstrate that ERIL resembles 
some modern imitation learning algorithms in terms of the number of
trajectories from expert data and outperformed sample efficiency
in terms of the number of trajectories in the forward RL step.
Ablation studies show that entropy regularization plays a critical
role in improving sample efficiency.
  Next we conducted a vision-based target-reaching task with
  a manipulator in three-dimensional space and
  demonstrated that using two discriminators is vital
  when the learner's initial state distribution differs
  from the expert one. 
Then we applied ERIL to human behaviors for performing a pole-balancing
task. Since the actions of human subjects are not observable, the task
is an example of realistic situations. ERIL recovers the subjects'
policies better than the baselines. We also showed that the estimated
reward functions show how every subject achieved her goal.

  The following are the main contributions of our paper:
  (1) We proposed a structured discriminator with hyperparameters
  derived from entropy-regularized reinforcement learning.
  (2) The hyperparameters, which are shared between forward and inverse
  RL, can be used to control the discriminator's ability.
  (3) The state value function is trained by both forward and
  inverse RL, which improves the sample efficiency in terms of the
  number of environmental interactions.

\section{Related Work}

\subsection{Regularization in RL}
\label{sec:rw:regularization}

The role of regularization is to prevent overfitting and to
encourage generalization. Many regularization methods
have been proposed from various aspects, such as
dropout \citep{Liu2021a},
temporal difference error \citep{Parisi2019a},
value function difference \citep{Ohnishi2019a}, and manifold
regularization for feature representation learning \citep{Li2018a}.
\citet{Amit2020a} investigated the property of a discount
factor as a regularizer.

The most widely used regularization method is entropy
regularization \citep{Ziebart2008a, Belousov2019a}.
It supports exploration by favoring more stochastic policies
\citep{Haarnoja2018a} and smoothens the optimization landscape
\citep{Ahmed2019a}. 
The advantage of entropy regularization in inverse RL is
its robustness against noisy and stochastic demonstrations
because an optimal policy becomes stochastic. 

\subsection{Behavior cloning}
\label{sec:rw:BC}

BC
directly maximizes the log-likelihood of an
expert action. \citet{Pomerleau1989a} achieved an Autonomous Land Vehicle
In a Neural Network (ALVINN), which is a potential precursor of
autonomous driving cars. ALVINN's policy is implemented by a 3-layer
neural network and learns mappings from video and range finder
inputs to steering directions by supervised learning. However, ALVINN suffers from the covariate shift problem.
To reduce covariate shift, \citet{Ross2011a} proposed an iterative
method called Dataset Aggregation (DAGGER), where the learner runs
its policy while the expert provides the correct action
for the states visited by the learner.
\citet{Laskey2017a} proposed Disturbances for Augmented Robot Trajectories
(DART), which collects expert datasets with injected noise. 

It is often difficult to provide such expert data as
state-action pairs in such realistic situations as analyzing animal
behaviors and learning from videos.
\citet{Torabi2018a} studied a situation in which expert action
is unavailable
and proposed Behavioral Cloning from Observation (BCO) that estimates
actions from an inverse dynamics model. 
Soft Q Imitation Learning (SQIL) \citep{Reddy2020a} is a BC with a
regularization term that penalizes large squared soft Bellman error.
SQIL is implemented by SAC, which assigns the reward of the expert
data to 1 and the generated data to 0.
SQIL can learn from both the expert and learner's samples
because SAC is an off-policy algorithm. 

\subsection{Generative adversarial imitation learning}
\label{sec:rw:GAIL}

GAIL \citep{Ho2016c}, which is a very popular imitation learning
algorithm, formulates the objectives of imitation learning
as GAN training objectives. Its discriminator 
differentiates between generated state-action pairs and those of experts, and
the generator acts as a forward reinforcement learning algorithm
to maximize the sum of rewards computed by the discriminator.
There are several extensions of GAN. AIRL \citep{Fu2018a} used an
optimal discriminator whose expert distribution was approximated
by a disentangled reward function.
A similar, independently proposed discriminator 
\citep{Uchibe2014b,Uchibe2018c} was derived from the framework
of entropy-regularized reinforcement learning and density ratio
estimation.
AIRL experimentally showed that the learned reward function can be transferred
to new, unseen environments.
Situated GAIL \citep{Kobayashi2019a} extended GAIL to learn multiple
reward functions and multiple policies by introducing a task variable
to both the discriminator and the generator.

As discussed in the previous section, expert action is not
always available. IRLGAN \citep{Henderson2018a} is a special GAN case
where the discriminator is given as a state function.
\citet{Torabi2019a} proposed Generative Adversarial Imitation from
Observation (GAIfO) whose functions are characterized by 
state transitions.
\citet{Sun2019a} proposed Action Guided Adversarial Imitation Learning
(AGAIL) that can deal with expert demonstrations with incomplete
action sequences. AGAIL uses mutual information between expert
and generated actions as an additional regularizer for
training objectives.

To reduce the number of interactions in the forward RL step,
\citet{Blonde2019a} proposed Sample-efficient Adversarial Mimic (SAM), which
adopts an off-policy method called the Deep Deterministic
Policy Gradients (DDPG) algorithm \citep{Lillicrap2016a}.
SAM maintains three different neural networks, which approximate
a reward function, a state-action value function, and a policy.
The reward function is estimated in the same way as in GAN, and the
state-action value function and the policy are trained by DDPG.
\citet{Kostrikov2019a} showed that GAIL's reward function is
biased and that the absorbing states are not treated appropriately.
They proposed a preprocessing technique for expert data before
learning and developed the Discriminator Actor-Critic (DAC) algorithm.
DAC utilizes the Twin Delayed Deep Deterministic policy gradient (TD3)
algorithm \citep{Fujimoto2018a}, which is an extension of DDPG.
\citet{Sasaki2019a} exploited the Bellman equation to represent a
reward function, and a reward's exponential transformation is
trained as a kind of discriminator. They improved the policy using an off-policy
actor-critic \citep{Degris2012a}.
\citet{Zuo2020a} proposed a Deterministic GAIL that adopts the modified
DDPG algorithm that incorporates the behavior cloning loss in the
forward RL step.
Discriminator Soft Actor-Critic \citep{Nishio2020a} extended SQIL by
estimating the reward function by an AIRL-like discriminator.
\citet{Ghasemipour2019a} and \citet{Ke2020a} described the relationship
between several imitation learning algorithms from the viewpoint of
objective functions.
Since our study focuses on sample efficiency with respect
to the number of interactions with the environment, the most related
works are SAM, DAC, and Sasaki's method. We compared our method with
these three methods in the experiments. 

\section{Preliminaries}
\label{sec:preliminaries}

\subsection{Markov decision process}
\label{sec:MDP}

Here we briefly introduce
MDP
for a discrete-time domain. 
Let $\mathcal{X}$ and $\mathcal{U}$ be continuous or discrete state and
action spaces. 
At time step $t$, a learning agent observes environmental current 
state $\bx_t \in \mathcal{X}$ and executes action 
$\bu_t \in \mathcal{U}$ sampled from stochastic policy 
$\pi(\bu_t \mid \bx_t)$. 
Consequently,
the learning agent receives from the environment immediate reward
  $\tilde{r}(\bx_t, \bu_t)$, which is an
  arbitrary bounded function that evaluates the goodness of
  action $\bu_t$ at state $\bx_t$.
The environment shifts to next state
  $\bx_t' = \bx_{t+1} \in \mathcal{X}$ according to state
  transition probability $p_T(\bx_t' \mid \bx_t, \bu_t)$.

Forward reinforcement learning's goal is to construct optimal policy
$\pi (\bu \mid \bx)$ that maximizes the given objective function.
Among several available objective functions, the most widely used is a 
discounted sum of rewards:
\begin{displaymath}
  V(\bx) \triangleq \mathbb{E} \left[ \sum_{t=0}^\infty \gamma^t 
  \tilde{r}(\bx_t, \bu_t) \mid \bx_0 = \bx \right],
\end{displaymath}
where $\gamma \in [0, 1)$ is called the discount factor. 
The optimal state value function for the discounted reward
setting satisfies the following Bellman optimality equation:
\begin{equation}
  \label{eq:MDP:BellmanV}
  V(\bx) = \max_{\bu} \left[ \tilde{r}(\bx, \bu) + \gamma 
  \mathbb{E}_{\bx' \sim p_T (\cdot \mid \bx, \bu)}
  \left[ V(\bx') \right] \right],
\end{equation}
where $\mathbb{E}_{p}[\cdot]$ hereafter denotes the expectation with respect to
$p$. 
The state-action value function for the discounted reward setting is also 
defined:
\begin{displaymath}
  Q(\bx, \bu) = \tilde{r}(\bx, \bu) + \gamma 
  \mathbb{E}_{\bx' \sim p_T (\cdot \mid \bx, \bu)}
  \left[ V(\bx') \right].
\end{displaymath}
Eq.~(\ref{eq:MDP:BellmanV}), which is nonlinear due to
the max operator, usually struggles to find an action that
maximizes its right hand side.

\subsection{Entropy-regularized Markov decision process}
\label{sec:ERL}

Next we consider entropy-regularized MDP
\citep{Azar2012a,Haarnoja2018a,Kozuno2019a,Ziebart2008a},
in which the reward function is regularized by the following form:
\begin{equation}
  \label{eq:ERMDP:reward}
  \tilde{r}(\bx, \bu) = r(\bx, \bu)
  + \frac{1}{\kappa} \mathcal{H}(\pi(\cdot \mid \bx))
  - \frac{1}{\eta}
  \mathrm{KL}(\pi(\cdot \mid \bx) \parallel b(\cdot \mid \bx)),
\end{equation}
where $r (\bx, \bu)$ is a standard reward function that is unknown in
the inverse RL setting.
$\kappa$ and $\eta$ are the positive hyperparameters determined by the
experimenter, $\mathcal{H}(\pi (\cdot \mid \bx))$ is the (differential)
entropy of policy $\pi (\bu \mid \bx)$, 
and $\mathrm{KL}(\pi(\cdot \mid \bx) \parallel b(\cdot \mid \bx))$ is the
relative entropy, which is also known as the
Kullback-Leibler (KL) divergence between $\pi (\bu \mid \bx)$ and baseline
policy $b(\bu \mid \bx)$. 
When the reward function is regularized by the entropy 
functions~(\ref{eq:ERMDP:reward}), we can analytically maximize the
right hand side of Eq.~(\ref{eq:MDP:BellmanV}) by a method using
Lagrange multipliers. 
Consequently, the optimal state value function can be represented:
\begin{equation}
  \label{eq:ERMDP:defV}
  V (\bx) = \frac{1}{\beta} \ln \int \exp \left( \beta Q(\bx, \bu)
  \right) \mathrm{d}\bu,
\end{equation}
where $\beta$ is a positive hyperparameter defined by
\begin{displaymath}
  \beta \triangleq \frac{\kappa \eta}{\kappa + \eta},
\end{displaymath}
and $Q(\bx, \bu)$ is the optimal soft state-action value function: 
\begin{equation}
  \label{eq:ERMDP:defQ}
  Q (\bx, \bu) = r(\bx, \bu) + \frac{1}{\eta} \ln b(\bu \mid \bx) + \gamma
  \mathbb{E}_{\bx' \sim p_T(\cdot \mid \bx, \bu)} 
  \left[ V (\bx') \right].
\end{equation}
When the action is discrete, the right hand side of 
Eq.~(\ref{eq:ERMDP:defV}) is a log-sum-exp function, also known
as a softmax function. 
The corresponding optimal policy is given:
\begin{equation}
  \label{eq:ERMDP:defpi}
  \pi (\bu \mid \bx) = \frac{\exp (\beta Q (\bx, \bu))}
  {\exp (\beta V(\bx))},
\end{equation}
where $\exp (\beta V (\bx))$ represents a normalizing
constant of $\pi (\bu \mid \bx)$.

For later reference, the update rules are described here.
The soft state-action value function is trained to minimize the soft
Bellman residual. The soft state value function is trained to minimize
the squared residual error derived from
Eq.~(\ref{eq:ERMDP:defpi}). 
The policy is improved by directly minimizing the expected
KL divergence in Eq.~(\ref{eq:ERMDP:defpi}):
\begin{displaymath}
  J_{\mathrm{ER}} (\bw_\pi) = \mathbb{E}_{\bx \sim p(\bx)} \left[
  \mathrm{KL} \left(
    \pi (\cdot \mid \bx) \parallel \frac{\exp (\beta Q(\bx, \cdot))}
    {\exp (\beta V(\bx))} 
  \right) \right].
\end{displaymath}
The derivative of the KL divergence is given by
\begin{equation}
  \label{eq:ERMDP:derivative_Jpi}
  \nabla \mathrm{KL} = \mathbb{E}_{\bu \sim \pi (\cdot \mid \bx)}
  \left[  
    \nabla_{\bw_\pi} \ln \pi (\cdot \mid \bx)
    \left[
      \ln \pi (\cdot \mid \bx) - \beta ( Q(\bx, \cdot) - V(\bx) ) 
      + B(\bx)
    \right]
 \right],
\end{equation}
where $B(\bx)$ is a baseline function that does not change the
gradient \citep{Peters2008a} and is often used to reduce the
variance of the gradient estimation. 

\subsection{Generative adversarial networks}
\label{sec:GAN}

GANs are a class of neural networks
that can approximate probability distribution based on a game
theoretic scenario \citep{Goodfellow2014a}.
Standard GANs consist of a generator and a discriminator.
  Suppose that $p^E(\bm{z})$ and $p^L(\bm{z})$ denote the
  probability distributions over data $\bm{z}$ of the
  expert and a generator.
A discriminator, which is a function that distinguishes samples from
a generator and an expert, is denoted by $D(\bm{z})$ and
minimizes the following negative log-likelihood
(NLL):
\begin{equation}
  \label{eq:GAN:discriminator_objective}
  J_{\mathrm{GAN}}^{(D)} = 
  - \mathbb{E}_{\bm{z} \sim p^L} [ \ln D(\bm{z}) ] -
  \mathbb{E}_{\bm{z} \sim p^E} [ \ln (1 - D(\bm{z})) ].
\end{equation}
The optimal discriminator has the following
shape \citep{Goodfellow2014a}:
\begin{equation}
  \label{eq:GAN:optimal_discriminator}
  D^\ast (\bm{z}) = \frac{p^L (\bm{z})}
  {p^L (\bm{z}) + p^E (\bm{z})}.
\end{equation}
The generator minimizes $- J_{\mathrm{GAN}}^{(D)}$: 
\begin{equation}
  \label{eq:GAN:generator_objective}
  J_{\mathrm{GAN}}^{(G)} = \mathbb{E}_{\bm{z} \sim p^L} [ \ln D(\bm{z}) ],
\end{equation}
where the second term on the right hand side of
Eq.~(\ref{eq:GAN:discriminator_objective}) is removed because it is
constant with respect to the generator.
Recently, Prescribed GAN (PresGAN) introduced an entropy
regularization term to $J_{\mathrm{GAN}}$ to mitigate mode
collapse \citep{Dieng2019a}.

GAIL \citep{Ho2016c} is an extension of GANs for imitation learning,
and its objective function is given by
\begin{align}
  J_{\mathrm{GAIL}}(\bw_G, \bw_D) &= 
  - \mathbb{E}_{(\bx, \bu) \sim \pi^L} [ \ln D(\bx, \bu) ] -
  \mathbb{E}_{(\bx, \bu) \sim \pi^E} [ \ln (1 - D(\bx, \bu)) ]
  \notag \\
  &\qquad + \lambda_{\mathrm{GAIL}} \mathcal{H}(\pi^L),
  \notag
\end{align}
where $\lambda_{\mathrm{GAIL}}$ is a positive hyperparameter.
Adding the entropy term is key for an association with the
entropy-regularized MDP. 
The objective function of the GAIL discriminator is essentially identical as that of the GAN discriminator, which
updates its policy by
  Trust Region Policy Optimization (TRPO) \citep{Schaul2015a},
where the reward function is defined by
\begin{displaymath}
  r_{\mathrm{GAIL}} = - \ln D(\bx, \bu).
\end{displaymath}

AIRL \citep{Fu2018a} adopts a special structure for the discriminator:
\begin{equation}
  \label{eq:AIRL:discriminator}
  D(\bx, \bu, \bx') = \frac{\pi^L (\bu \mid \bx)}
  {\exp (f(\bx, \bu, \bx')) + \pi^L (\bu \mid \bx)},
\end{equation}
where $f(\bx, \bu, \bx')$ is defined using two state-dependent
functions, $g(\bx)$ and $h(\bx)$:
\begin{equation}
  \label{eq:AIRL:f}
  f(\bx, \bu, \bx') \triangleq g(\bx) + \gamma h(\bx') - h(\bx). 
\end{equation}
Note that the AIRL discriminator~(\ref{eq:AIRL:discriminator}) has
no hyperparameters. A similar discriminator was proposed
\citep{Uchibe2014b,Uchibe2018c}.
AIRL's reward function is calculated:
\begin{align}
  r_{\mathrm{AIRL}} &= - \ln D(\bx, \bu) + \ln (1 - D(\bx, \bu))
  \notag \\
  &= f(\bx, \bu, \bx') - \ln \pi (\bu \mid \bx).
\end{align}
Note that $- \ln D$ and $\ln (1 - D)$ are monotonically related.

\subsection{Behavior cloning}
\label{sec:BC}

BC is the most widely used type of imitation learning.
It minimizes the ``forward'' KL divergence, which is also known as
M-projection \citep{Ghasemipour2019a,Ke2020a}:
\begin{equation}
  \label{eq:BC:objective}
  J_{\mathrm{BC}}(\bw_\pi) =
  \mathrm{KL}(\pi^E \parallel \pi^L) = - \mathbb{E}_{\pi^E}
  \left[ \ln \pi^L (\bu \mid \bx ) \right] + C,
\end{equation}
where $\bw_\pi$ is the policy parameter vector of $\pi^L(\bu \mid \bx)$
and $C$ is a constant with respect to the policy parameter.
Since BC does not need to interact with the environment, 
finding a policy that minimizes Eq.~(\ref{eq:BC:objective}) is
simply achieved by supervised learning.
Unfortunately, it suffers from covariate shift between the
learner and the expert \citep{Ross2011a}.

\section{Entropy-Regularized Imitation Learning}
\label{sec:ERIL}

\subsection{Objective function}
\label{sec:ERIL:objective}

To formally present our approach, we denote the 
expert's policy as $\pi^E (\bu \mid \bx)$ and the learner's policy as
$\pi^L (\bu \mid \bx)$. Suppose a dataset of transitions
generated by
\begin{displaymath}
  \mathcal{D}^E = \{ (\bx_i, \bu_i, \bx_i') \}_{i=1}^{N^E}, 
  \quad
  \bu_i \sim \pi^E (\cdot \mid \bx_i),
  \quad
  \bx_i' \sim p_T(\cdot \mid \bx_i, \bu_i),
\end{displaymath}
where $N^E$ is the number of transitions in the dataset. 
We consider two joint probability
density functions, $\pi^E(\bx, \bu, \bx')$ and $\pi^L(\bx, \bu, \bx')$,
where under a Markovian assumption, $\pi^E(\bx, \bu, \bx')$ is 
decomposed by
\begin{equation}
  \label{eq:ERIL:joint_prob_decomposition}
  \pi^E(\bx, \bu, \bx') = p_T(\bx' \mid \bx, \bu)
  \pi^E(\bu \mid \bx) \pi^E(\bx),
\end{equation}
and $\pi^L(\bx, \bu, \bx')$ can be decomposed in the same way. 

ERIL minimizes the ``reverse''
KL divergence given by
\begin{equation}
  \label{eq:ERIL:objective}
  J_{\mathrm{ERIL}} (\bw_\pi) = \mathrm{KL}(\pi^L \parallel \pi^E),
\end{equation}
where the reverse KL divergence is often called 
information projection (I-projection), defined by
\begin{displaymath}
  \mathrm{KL}(\pi^L \parallel \pi^E) = \mathbb{E}_{\pi^L} \left[
  \ln \frac{\pi^L(\bx, \bu, \bx')}{\pi^E(\bx, \bu, \bx')}
  \right].
\end{displaymath}
The difficulty is how to evaluate the log-ratio, 
$\ln \pi^L(\bx, \bu, \bx')/\pi^E (\bx, \bu, \bx')$, because
$\pi^E(\bx, \bu, \bx')$ is unknown. 
The basic idea for resolving the problem is to adopt the density ratio
trick \citep{Sugiyama2012b}, which can be efficiently achieved by
solving binary classification tasks. 
To derive the algorithm, we assume that the expert reward function
is given by
\begin{equation}
  \label{eq:ERIL:reward}
  \tilde{r}(\bx, \bu) = r_{k}(\bx)
  + \frac{1}{\kappa} \mathcal{H}(\pi^E (\cdot \mid \bx))
  - \frac{1}{\eta} \mathrm{KL}(\pi^E (\cdot \mid \bx) \parallel
  \pi_k^L (\cdot \mid \bx)),
\end{equation}
where $k$ is the iteration index.
$r_{k}(\bx)$ is a state-only reward function parameterized by
$\bw_r$. Although using a state-action reward function is possible,
the learned reward function will be a shaped advantage
function, and transferability to a new environment is
restricted \citep{Fu2018a}.
When the expert reward function is given by
Eq.~(\ref{eq:ERIL:reward}), the expert state-action value
function satisfies the following soft Bellman optimality equation:
\begin{equation}
  \label{eq:ERIL:Q}
  Q_k (\bx, \bu) = r_k(\bx) + \frac{1}{\eta} \ln \pi_k^L (\bu \mid \bx)
  + \gamma \mathbb{E}_{\bx' \sim p_T(\cdot \mid \bx, \bu)} 
  \left[ V_{k} (\bx') \right],
\end{equation}
  where $V_k(\bx’)$ is the corresponding state value function at
  the $k$-th iteration.
  Fig.~\ref{fig:eril_interaction} shows ERIL's overall architecture.
  The inverse RL step estimates reward $r_k$ and state value
  function $V_k$ from the expert's and learner's datasets.
  The forward RL step improves learner's policy $\pi_k^L$
  based on $r_k$ and $V_k$.
The state value function and the expert policy are expressed by
Eqs.~(\ref{eq:ERMDP:defV}) and Eq.~(\ref{eq:ERMDP:defpi}).

\subsection{Inverse reinforcement learning based on density ratio estimation}
\label{sec:IRL}

Arranging Eqs.~(\ref{eq:ERIL:Q}) and (\ref{eq:ERMDP:defpi}) yields 
the Bellman optimality equation in a different form:
\begin{equation}
  \label{eq:ERIL:Bellman1}
  \frac{1}{\beta} \ln 
  \frac{\pi^E (\bu \mid \bx)}{\pi^L_{k} (\bu \mid \bx)} =
  r_k (\bx) - \frac{1}{\kappa} \ln \pi^L_k (\bu \mid \bx)
  + \gamma \mathbb{E}_{\bx ' \sim p_T (\cdot \mid \bx, \bu)}
  \left[ V_k (\bx') \right] - V_k (\bx).
\end{equation}
With the density ratio trick \citep{Sugiyama2012b}, 
Eq.~(\ref{eq:ERIL:Bellman1}) is re-written:
\begin{equation}
  \label{eq:ERIL:Bellman2}
  \frac{1}{\beta}
  \ln \frac{D_k^{(2)} (\bx, \bu, \bx')}{1 - D_k^{(2)} (\bx, \bu, \bx')} =
  \frac{1}{\beta}
  \ln \frac{D_k^{(1)} (\bx)}{1 - D_k^{(1)} (\bx)} -
  r_k (\bx) + \frac{1}{\kappa} \ln \pi^L_k (\bu \mid \bx)
  - \gamma V_k (\bx') + V_k (\bx),
\end{equation}
where $D_k^{(1)}(\bx)$ and $D_k^{(2)}(\bx, \bu, \bx')$ denote a discriminator
that classifies the expert data from those of the generator. 
See \ref{sec:eq:ERIL:Bellman2} for the derivation.
Suppose that $\ln D_k^{(1)}(\bx) / (1 - D_k^{(1)}(\bx))$ is
approximated by $g_k (\bx)$ parameterized by $\bw_g$.
Then the first discriminator is constructed:
\begin{displaymath}
  D_k^{(1)} (\bx) = \frac{1}{1 + \exp (- g_k (\bx))},
\end{displaymath}
and $\bw_g$ is obtained by logistic regression.
In the same way, Eq.~(\ref{eq:ERIL:Bellman2}) is used to
design the second discriminator:
\begin{equation}
  \label{eq:ERIL:discriminator}
  D_k^{(2)} (\bx, \bu, \bx') =
  \frac{\exp (\beta \kappa^{-1} \ln \pi_k^L (\bu \mid \bx))}
  {\exp (\beta f_k(\bx, \bx')) + \exp (\beta \kappa^{-1} \ln \pi_k^L
  (\bu \mid \bx))},
\end{equation}
where
\begin{displaymath}
  f_k(\bx, \bx') \triangleq r_k(\bx) - \beta^{-1} g_k(\bx)
  + \gamma V_k(\bx') - V_k(\bx).
\end{displaymath}
Note that $V(\bx) = 0$ is required for absorbing states
$\bx \in \mathcal{X}$ so that the process can continue indefinitely
without incurring extra rewards. If $\bx'$ is an absorbing state, then 
$f_k(\bx, \bx') = r_k(\bx) + g_k(\bx) - V_k(\bx)$. 
When $V_k(\bx)$ is parameterized by $\bw_V$,
the parameters of $D^{(2)}$ are $\bw_r$ and $\bw_V$ because
$\bw_g$ and the policy are fixed during training.
Fig.~\ref{fig:eril_discriminator} shows the architecture of the
ERIL discriminator. 
The AIRL discriminator~(\ref{eq:AIRL:discriminator}) is a special case of
Eq.~(\ref{eq:ERIL:discriminator}) that sets
$g_k(\bx) = 0$ and $\beta \kappa^{-1} = 1$, where LogReg-IRL 
\citep{Uchibe2018c} is obtained by $\kappa \to \infty$. 
Note that the discriminator~(\ref{eq:ERIL:discriminator}) remains unchanged
even if $r_k(\bx)$ and $V_k(\bx)$ are modified by
$r_k(\bx) + C$ and $V_k (\bx) + C/(1 - \gamma)$, where $C$ is a constant
value. Therefore, we can recover them up to the constant. 

Our inverse RL step consists of two parts. First, $g_k(\bx)$ is evaluated 
by maximizing the following log-likelihood:
\begin{equation}
  \label{eq:ERIL:loss_D1}
  J_{D}^{(1)} (\bw_g) = \mathbb{E}_{\bx \sim \mathcal{D}_k^L} \left[
  \ln D_k^{(1)}(\bx) \right] + 
  \mathbb{E}_{\bx \sim \mathcal{D}^E} \left[
  \ln \left( 1 - D_k^{(1)} (\bx) \right) \right],
\end{equation}
where $\bx \sim \mathcal{D}$ denotes that the transition
data are drawn from $\mathcal{D}$ and 
$\mathcal{D}^L_k$ is the dataset of the transitions
provided by learner's policy $\pi^L_k (\bu \mid \bx)$ at the $k$-th
iteration. 
$D_k^{(2)}(\bx, \bu, \bx')$ is
trained in the same way, in which $g_k(\bx)$ and $\pi^L_k(\bu \mid \bx)$
are fixed during training. The goal of
$D_k^{(2)}(\bx, \bu, \bx')$ is to maximize the following log-likelihood:
\begin{equation}
  \label{eq:ERIL:loss_D2}
  J_{D}^{(2)}(\bw_r, \bw_V)
  =
  \mathbb{E}_{(\bx, \bu, \bx') \sim \mathcal{D}_k^L} \left[
  \ln D_k^{(2)}(\bx, \bu, \bx') \right] + 
  \mathbb{E}_{(\bx, \bu, \bx') \sim \mathcal{D}^E} \left[
  \ln \left( 1 - D_k^{(2)} (\bx, \bu, \bx') \right) \right].
\end{equation}
This simply minimizes the cross entropy function of 
binary classifier $D_k^{(2)}(\bx, \bu, \bx')$ that separates the 
generated data from those of the expert.
The formulation of the inverse RL algorithm in ERIL is identical
to that of the GAN discriminator.
Note that $\beta$ is not estimated as an independent parameter.

In practice, $\mathcal{D}_k^L$ is replaced with
$\mathcal{D}^L = \cup_k \mathcal{D}_k^L$, which means that the
discriminators are trained with all the generated transitions
by the learner. Theoretically, we have to use importance sampling
to correctly evaluate the expectation, but
\cite{Kostrikov2019a} showed that it works well without
it.

\subsection{Forward reinforcement learning based on KL minimization}
\label{sec:FRL}

After estimating the log-ratio by the inverse RL described in
Section~\ref{sec:IRL}, we minimize the reverse KL
divergence~(\ref{eq:ERIL:objective}).
Then we rewrite Eq.~(\ref{eq:ERIL:Bellman2}) using
Eq.~(\ref{eq:ERIL:Q}) and obtain the following equation:
\begin{equation}
  \label{eq:ERIL:Bellman3}
  \ln \frac{D_k^{(2)} (\bx, \bu, \bx')}{1 - D_k^{(2)} (\bx, \bu, \bx')}
  =
  \ln \pi_k^L (\bu \mid \bx) - \beta \left( Q_k (\bx, \bu)
    - V_k(\bx) \right) + g_k (\bx).
\end{equation}
ERIL's objective function is expressed by
\begin{equation}
  \label{eq:ERIL:approximate_objective}
  J_{\mathrm{ERIL}} (\bw_\pi) = \mathbb{E}_{\pi^L} \left[
  \ln \pi_k^L (\bu \mid \bx) - \beta \left( Q_k (\bx, \bu)
    - V_k(\bx) \right) + g_k (\bx)
  \right]
\end{equation}
and its derivative by
\begin{equation}
  \label{eq:ERIL:derivative_Jpi}
  \nabla J_{\mathrm{ERIL}}(\bw_\pi) = \mathbb{E}_{\pi^L} \left[
  \nabla_{\bw_\pi} \ln \pi_k^L (\bu \mid \bx) \left[
   \ln \pi_k^L (\bu \mid \bx) - \beta ( Q_k(\bx, \bu) - V_k(\bx) ) 
   + g_k(\bx) \right]
 \right].
\end{equation}
When $g_k(\bx) = B(\bx)$, 
Eq.~(\ref{eq:ERIL:derivative_Jpi}) is essentially identical to
Eq.~(\ref{eq:ERMDP:derivative_Jpi}). 

Our forward RL step updates $V_k$ and $Q_k$ like the standard
SAC algorithm \citep{Haarnoja2018a}.
The loss function of the state value function is given:
\begin{equation}
  \label{eq:ERIL:loss_V}
  J_{\mathrm{ERIL}}^{V} (\bw_V) = \frac{1}{2}
  \mathbb{E}_{\bx \sim \mathcal{D}}
  \left[ \left( V(\bx) - \mathbb{E}_{\bu \sim \pi_k^L (\cdot \mid \bx)}
      \left[ Q_k (\bx, \cdot) - \frac{1}{\beta} \ln \pi_k^L (\cdot \mid \bx)
      \right] \right)^2
  \right].
\end{equation}
The state-action value function is trained to minimize the soft Bellman residual
\begin{equation}
  \label{eq:ERIL:loss_Q}
  J_{\mathrm{ERIL}}^{Q} (\bw_Q) = \frac{1}{2}
  \mathbb{E}_{(\bx, \bu, \bx') \sim \mathcal{D}^L} \left[
    \left( Q(\bx, \bu) - \bar{Q}_k (\bx, \bu, \bx') \right)^2
  \right],
\end{equation}
with
\begin{displaymath}
  \bar{Q}_k (\bx, \bu, \bx') = r_k(\bx) + \frac{1}{\eta} \ln
  \pi_k^L (\bu \mid \bx) + \gamma \bar{V}_k(\bx'),
\end{displaymath}
where $\tilde{V}_k(\bx)$ is the target state value function
parameterized by $\bar{\bw}_V$.
  Since our method is a model-free approach, we cannot compute
  the expected value of Eq.~(\ref{eq:ERMDP:defQ}).
  Instead, we use $\bar{Q}(\bx, \bu, \bx’)$, which is an
  approximation of $Q(\bx, \bu)$, and
  $Q(\bx, \bu) = \mathbb{E}_{\bx’ \sim p_T(\cdot \mid \bx, \bu)}[\bar{Q}(\bx, \bu, \bx’)]$.
Two alternatives can be chosen to update $\bar{\bw}_V$.
The first is a periodic update, i.e., where the target network is
synchronized with current $\bw_g$ at regular intervals
\citep{Mnih2015a}.
We use the second alternative, which is a Polyak averaging update,
where $\bar{\bw}_g$ is updated by a weighted average over the past
parameters \citep{Lillicrap2016a}:
\begin{equation}
  \label{eq:ERIL:update_target}
  \bar{\bw}_V \leftarrow \tau \bw_V + (1 - \tau) \bar{\bw}_V, 
\end{equation}
where $\tau \ll 1$. 

Algorithm~\ref{alg:ERIL} shows an overview of Entropy-Regularized
Imitation Learning. Lines 4-5 and 6-8 represent the inverse RL 
and forward RL steps.
Note that $\bw_V$ is updated twice in each iteration. 
  Since the second discriminator depends on the first one,
  update $\bw_g$ first, followed by $\bw_r$ and
  $\bw_V$.
  The order of Lines 6-8 is exchangeable in practice.

\begin{algorithm}[t]
\caption{Entropy-Regularized Imitation Learning}
\label{alg:ERIL}
\begin{algorithmic}[1]
  \Require{Expert dataset $\mathcal{D}^E$ and hyperparameters
    $\kappa$ and $\eta$}
  \Ensure{Learner's policy $\pi^L$ and estimated reward $r(\bx)$.} 
  \State Initialize all parameters of networks and replay buffer
  $\mathcal{D}^L$. 
  \For{$k = 0, 1, 2, \ldots$}
  \State Sample $\tau_i = \{ (\bx_t, \bu_t, \bx_{t+1} ) \}_{t=0}^T$
  with $\pi_k^L$ and store it in $\mathcal{D}^L$.
  \State Update $\bw_g$ with the gradient of Eq.~(\ref{eq:ERIL:loss_D1}).
  \State Update $\bw_r$ and $\bw_V$ with the gradient of
  Eq.~(\ref{eq:ERIL:loss_D2}).
  \State Update $\bw_Q$ with the gradient of Eq.~(\ref{eq:ERIL:loss_Q}).
  \State Update $\bw_V$ with the gradient of Eq.~(\ref{eq:ERIL:loss_V}).
  \State Update $\bw_\pi$ with the gradient of
  Eq.~(\ref{eq:ERIL:approximate_objective}).
  \State Update $\bar{\bw}_V$ by Eq.~(\ref{eq:ERIL:update_target}). 
  \EndFor
\end{algorithmic}
\end{algorithm}

\subsection{Interpretation of second discriminator}
\label{sec:interpretation_D2}

We show the connection between $D^{(2)}$ and the optimal discriminator
of GAN shown in Eq.~(\ref{eq:GAN:optimal_discriminator}).
Arranging Eq.~(\ref{eq:ERIL:Bellman3}) and assuming
$\lvert \mathcal{D}^E \rvert = \lvert \mathcal{D}^L \rvert$
yield the second discriminator:
\begin{equation}
  \label{eq:ERIL:optimal_discriminator}
  D^{(2)}(\bx, \bu) =
  \frac{\pi^L(\bx) \pi^L(\bu \mid \bx)}
  {\pi^E (\bx) \tilde{\pi}^E (\bu \mid \bx) + \pi^L(\bx)
    \pi^L (\bu \mid \bx)},
\end{equation}
where $\tilde{\pi}^E (\bu \mid \bx) \triangleq \exp (\beta (Q(\bx, \bu) - V(\bx))$.
We omit $\bx'$ from the input to $D^{(2)}$ here because it does not
depend on the next state.
See \ref{sec:eq:ERIL:optimal_discriminator} for the derivation.
By comparing Eqs.~(\ref{eq:GAN:optimal_discriminator}) and
(\ref{eq:ERIL:optimal_discriminator}), $D^{(2)}$ represents the
form of the optimal discriminator, and the expert policy is approximated
by the value functions. 

\subsection{Extension}
\label{sec:extendedERIL}

Here we describe two extensions to deal with more realistic situations.
One learns multiple policies from multiple experts.
The dataset of experts is augmented by adding conditioning variable
$c$ that represents the index of experts:
\begin{displaymath}
  \mathcal{D}^E = \{ (\bx_i, \bu_i, \bx_i', c) \}_{i=1}^{N^E}, 
  \quad
  \bu_i \sim \pi^E (\cdot \mid \bx_i, c),
  \quad
  \bx_i' \sim p_T(\cdot \mid \bx_i, \bu_i),
\end{displaymath}
where $c$ is often encoded as a one-hot vector.
Then we introduce universal value function approximators
\citep{Schaul2015a} to extend the value functions to be conditioned
on the subject index.
For example, the second discriminator is extended:
\begin{displaymath}
  D_k^{(2)} (\bx, \bu, \bx', c) = \frac{\exp (\beta f_k(\bx, \bx', c))}
  {\exp (\beta f_k(\bx, \bx', c)) + \exp (\beta \kappa^{-1} \ln \pi_k^L
  (\bu \mid \bx, c))},
\end{displaymath}
where $f_k(\bx, \bx', c) \triangleq r_k(\bx, c) + \beta^{-1} g_k(\bx, c) + \gamma V_k(\bx', c) - V_k(\bx, c)$.

The other extension deals with the case where actions are not 
observed. ERIL needs to observe the expert action because
$\mathcal{D}^{(2)}$ explicitly depends on the action.
One simple solution is to set $\kappa^{-1} = 0$. This is the special case
of ERIL in which the inverse RL step is LogReg-IRL
\citep{Uchibe2014b,Uchibe2018c}.
An alternative is to exploit an inverse dynamics model \citep{Torabi2018a}, which
is formulated as a maximum-likelihood estimation problem by
maximizing the following function:
\begin{displaymath}
  J_{\mathrm{IDM}}(\bw_a) = \sum_{(\bx_t, \bu_t, \bx_{t+1}) \in
    \mathcal{D}^L} \ln p(\bu_t \mid \bx_t, \bx_{t+1}),
\end{displaymath}
where $\bw_a$ is a parameter of the conditional probability density
over actions given a specific state transition.
Then the expert dataset is augmented:
\begin{displaymath}
  \tilde{\mathcal{D}}^E = \{ (\bx_i, \tilde{\bu}_i, \bx_i', c) \}_{i=1}^{N^E}, 
  \quad
  \tilde{\bu}_i \sim p (\cdot \mid \bx_i, \bx_i').
\end{displaymath}
Using these inferred actions, we apply ERIL as usual. 

\section{MuJoCo Benchmark Control Tasks}
\label{sec:mujoco}

\subsection{Task description}

We evaluated ERIL with six MuJoCo-simulated \citep{Todorov2012a}
environments, 
provided by the OpenAI gym \citep{Brockman2016a}: Hopper, Walker, Reacher, Half-Cheetah, Ant, and Humanoid.
The goal for all the tasks was to move forward as quickly as possible.
First, optimal policy $\pi^E (\bu \mid \bx)$ for every task was
trained by TRPO 
with a reward function provided by the OpenAI gym, and expert 
dataset $\mathcal{D}^E$ was created by executing $\pi^E$. 
Then we evaluated the imitation performance with respect to the sample
complexity of the expert data by changing the number of samples in
$\mathcal{D}^E$ and the number of interactions with the environment. 
Based on \citet{Ho2016c}, the trajectories constituting
each dataset consisted of about 50 transitions
$(\bx, \bu, \bx')$.
The same demonstration data were used to train all the algorithms for
a fair comparison.
We compared ERIL with BC, GAIL, Sasaki, DAC, and SAM. 
The network architectures that approximate the functions are
shown in Table~\ref{table:mujoco:networks}.
We used a Rectified Linear Unit (ReLU) activation function in the
hidden layers. The output nodes used a linear activation function, except $\mu(\bx)$ and $\sigma(\bx)$. 
Functions $\mu(\bx)$ and $\sigma(\bx)$ represent
the learner's policy by a Gaussian distribution:
\begin{displaymath}
  \pi^L (\bu \mid \bx) = \mathcal{N}(\bu \mid \mu(\bx), \sigma(\bx)),
\end{displaymath}
where $\mu(\bx)$ and $\sigma(\bx)$ denote the mean and the diagonal
covariance matrix.
The output nodes of $\mu(\bx)$ and $\sigma(\bx)$ use
tanh and sigmoid functions. 

\begin{table}[t]
  \centering
  \caption{Neural network architectures used in MuJoCo benchmark
    control tasks: $V(\bx)$ is approximated by a two-layer
    fully-connected neural network consisting of (256, 256) hidden
    units in HalfCheetah and Humanoid tasks.}
  \label{table:mujoco:networks}
  \begin{tabular}{lll} \hline
    Function & Number of nodes & Task \\ \hline \hline
    $\mu(\bx)$ & ($\dim \mathcal{X}$, 256, 256, $\dim \mathcal{U}$)
                               & HalfCheetah and Humanoid
    \\ \cline{2-3}
             & ($\dim \mathcal{X}$, 100, 100, $\dim \mathcal{U}$)
                               & Other tasks \\ \hline
    $\sigma(\bx)$ & ($\dim \mathcal{X}$, 100, $\dim \mathcal{U}$)
                               &All tasks \\ \hline
    $r(\bx)$ & ($\dim \mathcal{X}$, 100, 100, 1) & All tasks
    \\ \hline
    $V(\bx)$ & ($\dim \mathcal{X}$, 256, 256, 1) & HalfCheetah and Humanoid
    \\ \cline{2-3}
             & ($\dim \mathcal{X}$, 100, 100, 1) & Other tasks \\ \hline
    $Q(\bx, \bu)$ & ($\dim \mathcal{X} + \dim \mathcal{U}$,
                    256, 256, 1) & HalfCheetah and Humanoid
    \\ \cline{2-3}
             & ($\dim \mathcal{X} + \dim \mathcal{U}$,
               100, 100, 1) & Other tasks \\ \hline
    $g(\bx)$ & ($\dim \mathcal{X}$, 100, 100, 1) & All tasks
    \\ \hline
    $D(\bx, \bu)$ & ($\dim \mathcal{X} + \dim \mathcal{U}$,
                    256, 256, 1) & HalfCheetah and Humanoid
    \\ \cline{2-3}
             & ($\dim \mathcal{X} + \dim \mathcal{U}$,
               100, 100, 1) & Other tasks \\ \hline
  \end{tabular}
\end{table}

The number of trajectories generated by $\pi^L_k (\bu \mid \bx)$ was set 
to 100. 
In all of our experiments, the hyperparameters for regularizing the rewards were
$\kappa = 1$ and $\eta = 10$, and the discount factor was $\gamma = 0.99$
.
We trained all the networks with the Adam optimizer \citep{Kingma2015a}
and a decay learning rate. 
Following \citep{Fujimoto2018a}, we performed evaluations using ten
different random seeds. 

\subsection{Comparative evaluations}

Figure~\ref{fig:mujoco:frl_normal} shows the normalized
return of the evaluation rollouts during training for ERIL, BC, GAIL, Sasaki,
DAC, and SAM. A normalized return is defined by
\begin{displaymath}
  \bar{R} = \frac{R - R_0}{R^E - R_0},
\end{displaymath}
where $R$, $R_0$, and $R^E$ respectively denote the total returns of the learner,
a randomly initialized Gaussian policy, and an
expert trained by TRPO.
The horizontal axis depicts the number of interactions with the
environment in the logarithmic scale.
ERIL, Sasaki, DAC, and SAM have considerably
better sample efficiency than BC and GAIL in all the MuJoCo control
tasks. In particular, ERIL performed consistently across all the tasks
and outperformed the baseline methods in each one.
DAC outperformed Sasaki and SAM in Walker2d,
HalfCheetah, and Ant, and its performance was competitive with that of Hopper,
Reacher, and Humanoid. Sasaki achieved better performance than
SAM in Walker2D, and there was no significant difference between
Sasaki and SAM in our implementation. GAIL was not very
competitive because the on-policy TRPO algorithm used in its
forward RL step could not use the technique of experience replay.

Figure~\ref{fig:mujoco:irl_normal} shows the normalized
return of the evaluation rollout of the six algorithms with respect to
the number of demonstrations. ERIL, Sasaki,
DAC, and SAM were more sample-efficient than BC and GAIL. 
ERIL was consistently one of the most sample-efficient
algorithms when the number of demonstrations was small. 
DAC's performance was comparable to that of ERIL.
SAM showed comparable performance to ERIL in Hopper, but its
normalized return was worse than ERIL when the number of demonstrations
was limited. 
Although the BC implemented in this paper did not consider the covariate
shift problem, the normalized return approached 1.0 with 25
demonstrations. Note that BC is an M-projection that avoids
$\pi^L = 0$ whenever $\pi^E > 0$. Therefore, the policy trained by BC is
averaged over several modes, even if it is approximated
by a Gaussian distribution. On the contrary, ERIL is an
I-projection that forces $\pi^L$ to be zero even if $\pi^E > 0$.
Although the policy trained by ERIL concentrates on a single mode, BC obtained a comparable policy to ERIL because
the expert policy obtained by TRPO was also approximated by the
Gaussian distribution.

\subsection{Ablation study}

Next we investigate which ERIL component contributed most to its
performance by an ablation study on the Ant environment.
We tested five different components:
\begin{enumerate}
\item ERIL without the first discriminator, i.e., we set
  $g_k(\bx) = 0$ for all $\bx$.
\item ERIL without sharing the state value function between the forward
  and inverse RLs. The forward RL step updates the policy with 
  reward $r_k(\bx)$.
\item ERIL without sharing the state value function between the forward
  and inverse RL. The forward RL step updates the policy with 
  shaping reward $r_k(\bx) + \gamma V_k(\bx') - V_k(\bx)$ for 
  state transition $(\bx, \bx')$.
\item ERIL without sharing the hyperparameters.
  We set $\beta = 0$ and $\kappa = 0$ in
  Eq.~(\ref{eq:ERIL:discriminator}), and the $\beta$ and $\eta$ of 
  the forward RL step are used as they are.
\item ERIL without the soft Bellman equation. Discriminator 
  $D_k^{(3)}(\bx, \bu, \bx')$ is defined by
  \begin{displaymath}
    D_k^{(3)}(\bx, \bu, \bx') = \frac{1}{1 + \exp (-h_k(\bx, \bu, \bx'))},
  \end{displaymath}
  where $h_k(\bx, \bu, \bx')$ is directly implemented by a neural
  network. Then the reward function is computed by
  $r = -\ln D_k^{(3)}(\bx, \bu, \bx') + \ln (1 - D_k^{(3)}(\bx, \bu, \bx'))$
  like AIRL. 
\end{enumerate}

Figure~\ref{fig:mujoco:frl_ablation} compares the learning curves. 
In the benchmark tasks, we found no significant difference between 
the original ERIL and the ERIL without the first discriminator. 
One possible reason is that $g_k(\bx)$ did not change the policy 
gradient, as we explained in Section~\ref{sec:FRL}.
The other reason might be that $g_k(\bx)$ was close to zero because the
state distributions of the experts and learners did not differ
significantly due to the small variation of the initial states.
Regarding the property of sharing the state value function,
the results depended on how the reward was calculated from the
results of the inverse RL step. When the policy was trained with
$r_k(\bx)$, it took longer steps to reach the performance of the
original ERIL. On the other hand, if the reward is calculated by the
form of the shaping reward, there was no significant difference 
compared with the original ERIL.
When we removed the hyperparameters from the second discriminator,
slower learning resulted. The ERIL without the soft Bellman
equation was the most inefficient sample in the early stage of
learning for several reasons. One is that the 
discriminator is larger than the original ERIL because
the it was defined in the joint space of the state, the action,
and the next state. As a result, it required more
samples for training the discriminator. The second reason is that
there were no hyperparameters.

\section{Real Robot Experiment}
\label{sec:nextage}

  We further investigated the role of the first discriminator by
  conducting a robot experiment in which the 
  learner's initial distribution differs from the expert's
  initial distribution.

\subsection{Task description}
\label{sec:nextage:setting}

  We performed a reaching task defined as controlling the
  end-effector to a target position. We used an upper-body
  robot called Nextage developed and manufactured by Kawada
  Industries, Inc.(Fig.~\ref{fig:nextage:task}(a)).
  Nextage has a head with two cameras, a torso, two 6-axis
  manipulators, and two cameras attached to its end-effectors.
  We used the left arm, a camera mounted on it,
  and the left camera on the head in this task.
  The head pose was fixed during the experiments.
  There were two colored blocks as obstacles and one metal can that
  indicated the goal position in the workspace.
  We prepared several environmental configurations by changing
  the arm's initial pose, the can's location, and
  the height of the blocks.
  To evaluate the effect of the first discriminator, we
  considered the six initial poses shown in
  Fig.~\ref{fig:nextage:task}(b).
  The expert agent controlled the arm with three different
  initial poses, and the learning agent controlled it with five
  different initial poses, including the expert's initial
  poses. One pose was used to evaluate the learned policies.
  We prepared blocks of two different heights.
  Three goal positions were set: two for training
  and one for testing. The environmental
  configurations were constructed as follows:
\begin{itemize}
\item Expert configuration: An initial pose and a target
  pose were selected from three possible poses
  (black-filled circle) and two possible poses (black-filled squares),
  shown in Fig.~\ref{fig:nextage:task}(b).
  We randomly chose the short blocks and the tall ones. Consequently,
  there were $3 \times 2 \times 2 = 12$ configurations.
\item Learner's configuration: An initial pose and a target
  pose were selected from five possible poses
  (black-filled circle and black-empty circle) and two possible poses
  (black-filled square). We randomly chose short blocks and tall
  ones. Consequently,
  there were $5 \times 2 \times 2 = 20$ configurations.
  Note that eight configurations were not included in the
  expert configuration.
\item Test configuration: An initial pose was the red-filled
  circle shown in Fig.~\ref{fig:nextage:task}(b).
  The red-filled square represents the target position. 
  There were $1 \times 2 \times 2 = 4$ configurations.
\end{itemize}

  We used MoveIt! \citep{Chitta2012a}, which is the most
  widely used software for motion planning, to create
  expert demonstrations using the geometric information of
  the metal can and the colored blocks.
  Such information was not available for learning the
  algorithms. MoveIt! generated 20 trajectories for every
  expert configuration.
  We recorded the RGB images and joint angles.
  The sequence of the joint angles was almost deterministic,
  but that of the RGB images was stochastic and noisy due to the
  lighting conditions.

  The state of this task consists of six joint angles,
  $\{ \theta_i \}_{i=1}^6$, of the left arm and two
  $84 \times 84$ RGB images, $I_{\mathrm{head}}$ and 
  $I_{\mathrm{hand}}$, captured by Nextage's cameras.
  The action is given by the changes in the joint angle from
  previous position $\{ \Delta \theta_i \}_{i=1}^6$,
  where $\Delta \theta_i$ is the change in the joint angle from
  the previous position of the $i$-th joint.
  We provided a deep convolutional encoder to find an
  appropriate representation from the pixels based on a
  previous study \citep{Yarats2020a}.
  Fig.~\ref{fig:nextage:network}(a) shows the relationship
  among the networks used in the experiment. The policy, the reward,
  the state value, the state-action value, and the first discriminator
  shared the encoder network.
  The encoder maps captured images $I_{\mathrm{head}}$ and
  $I_{\mathrm{hand}}$ to latent variable $\bm{z}$:
  $\mathrm{enc}(I_{\mathrm{head}}, I_{\mathrm{hand}}) = \bm{z}$.
  Then it was concatenated with the joint angles for the
  input of the first discriminator, the reward, and the state value
  function. The action vector is also added to the policy
  and the state-action value function.
  The decoder is introduced to train the encoder.
  Fig. 7(b) shows the network architecture of the encoder
  network.

  Based on a suggestion inferred from research by \citet{Yarats2020a},
  we used a deterministic Regularized Autoencoder (RAE)
  \citep{Ghosh2020a}, which imposed an L2 penalty on
  learned representation $\bm{z}$ and a weight-decay on the
  decoder parameters.
  We prevented the gradients of the policy and the
  state-action value function from updating the convolutional
  encoder, another idea borrowed from 
  \citet{Yarats2020a}.
  See \ref{sec:RAE} for details.

\subsection{Experimental results}
\label{sec:nextage:results}

  We compared the normal ERIL with (1) the ERIL without the first
  discriminator, (2) DAC, (3) AIRL, and (4) BC.
  To measure the performances, we considered a synthetic reward
  function given by
  \begin{displaymath}
    r = \exp \left( - \frac{\| \bm{p}_t - \bm{p}_g \|_2^2}{\sigma^2}
      \right),
  \end{displaymath}
  where $\bm{p}_t$ and $\bm{p}_g$ denote the end-effector's current and the
  target position. Parameter $\sigma$
  was set to 5. Note that TRPO could not find an optimal policy
  from the synthetic reward due to the sparseness of the reward.

  The averaged learning results based on the three experiments
  are shown in Fig.~\ref{fig:nextage:performance}.
  The most sample-efficient method was ERIL, which achieved
  the best asymptotic performance.
  ERIL without the first discriminator and DAC learned
  efficiently at the early stage of learning, but their
  asymptotic performances were worse than that of ERIL.
  AIRL also achieved the same performance as ERIL without
  the first discriminator and DAC at the end of learning,
  and BC achieved the worse performance. 

  To evaluate the learned policies, we computed NLL for the
  trajectories generated by MoveIt!.
  Fig.~\ref{fig:nextage:nll} compares the NLL for the learning
  configuration (starting from two positions not included in
  the expert configuration) and those for the test configuration.
  Note that $g(\bx) \neq 0$ for these initial positions in
  the learning configuration.
  ERIL obtained a smaller NLL than other methods for both
  the configurations, suggesting  that the policy obtained by
  ERIL was closer to that of MoveIt!.
  We found that ERIL assigned a small reward value close to
  zero around these positions because the first discriminator
  was more dominant than the second one.
  These results suggest that the first discriminator plays
  a role when the learner's initial distribution differs
  from the expert one. 

\section{Human Behavior Analysis}
\label{sec:human_balancing}

\subsection{Task description}
\label{sec:human_balancing_task}

Next we evaluated ERIL in a realistic situation by conducting
a dynamic motor control experiment in which a human subject solved a
planar pole-balancing problem.
Fig.~\ref{fig:human:task}(a) shows the experimental setup.
The subject can move the base in the left, right, top, and bottom directions
to swing the pole several times and decelerate it to 
balance it in the upright position.
As shown in Fig.~\ref{fig:human:task}(b), 
the dynamics are described by a six-dimensional state vector,
$\bm{x} = [\theta, \dot{\theta}, x, \dot{x}, z, \dot{z}]^\top$, 
where $\theta$ and $\dot{\theta}$ are the angle and the angular velocity of
the pole, $x$ and $z$ are the horizontal and vertical positions of
the base, and
$\dot{x}$ and $\dot{z}$ are their time derivatives.
The state variables are defined in the following ranges: 
$\theta \in [-\pi, \pi]$ (in [rad]), $\dot{\theta} \in [-4\pi, 4\pi]$ (in 
[rad/s]), $x \in [0, 639]$ (in [pixel]), $\dot{x} \in [-5, 5]$ (in [pixel/s]), 
$z \in [0, 319]$ (in [pixel]), and $\dot{z} \in [-5, 5]$ (in [pixel/s]). 
Note that no applied forces $F_x$ and $F_z$ were observed on the pendulum in 
Fig.~\ref{fig:human:task} in this 
experiment. 

The task was performed under two pole conditions: long (73 cm) and
short (29 cm). Each subject had 15 trials to balance the pole in
each condition after some practice. 
Each trial ended when the subject could keep the pole upright for
three seconds or after 40 seconds had elapsed.  
We collected data from seven subjects (five right-handed and
two left-handed). We used a trajectory-based sampling method 
to construct the following three expert datasets:
$\mathcal{D}_{i, j, \mathrm{tr}}^E$ for training, 
$\mathcal{D}_{i, j, \mathrm{va}}^E$ for validation, 
and $\mathcal{D}_{i, j, \mathrm{te}}^E$ for testing the $i$-th subject. 
Subscript $j$ indicates 1 for the long-pole conditions and
2 for the short-pole conditions.

Since we had multiple experts whose actions were 
unavailable, we used the extended ERIL described in
Section~\ref{sec:extendedERIL}. We augmented the ERIL functions by a seven-dimensional conditional one-hot vector $\bm{c}$ 
and evaluated two ERIL variations. One is where the second 
discriminator was replaced by the LogReg-IRL by setting 
$\kappa^{-1} = 0$. This method is called ERIL($\kappa^{-1} = 0$). 
The other is the ERIL with IDM, in which the expert action is 
estimated by the inverse dynamics model. The two ERILs were compared 
with GAIfO, IRLGAN, and BCO. Since the two ERILs, GAIfO, and IRLGAN 
improved the policy by forward RL, they require a simulator of 
the environment. We modeled the dynamics shown in 
Fig.~\ref{fig:human:task}(b) as an X-Z inverted pendulum
\citep{Wang2012a}.
  The physical parameters and the motion of the equations are provided
  in \ref{sec:xz_inverted_pendulum}.

We parameterized the log of the reward function as a quadratic
function of the nonlinear features of the state:
\begin{displaymath}
  \ln r_k(\bx, c) = - \frac{1}{2} (\bx - \bm{m}(\bx, \bm{c}))^\top 
  \bm{P}(\bm{x}, \bm{c}) (\bx - \bm{m}(\bx, \bm{c})),
\end{displaymath}
where $\bm{c}$ is a vector encoding the subject index and condition.
$\bm{c}$ is a concatenation of $\bm{c}^s$ and
$\bm{c}^c$ that denotes the one-hot vector to encode the subject and
the condition. Since we had seven subjects and two
experimental conditions, $\bm{c}$ was a seven-dimensional vector.
$\bm{P}(\bx, \bm{c})$ is a positive definite square matrix,
parameterized by
$\bm{P}(\bx, \bm{c}) = \bm{L}(\bx, \bm{c}) \bm{L}(\bx, \bm{c})^\top$,
where $\bm{L}(\bx, \bm{c})$ is a lower triangular matrix. 
The $\bm{L}(\bm{x}, \bm{c})$ element is a linear output layer of
the neural network with exponentially transformed diagonal terms. 
Table~\ref{table:human:networks} shows the network architecture.
As discriminators, note that IRLGAN uses $D(\bx, \bm{c})$, and GAIfO uses
$D(\bx, \bx', \bm{c})$. 

\begin{table}[t]
  \centering
  \caption{Neural network architectures used in pole-balancing task:
    For example, $V(\bx)$ is approximated by a two-layer
    fully-connected neural network consisting of (256, 256) hidden
    units in HalfCheetah and Humanoid tasks.}
  \label{table:human:networks}
  \begin{tabular}{ll} \hline
    Function & Number of nodes \\ \hline \hline
    $\mu(\bx, \bm{c})$ & ($\dim \mathcal{X} + \dim \bm{c}$, 100,
                         $\dim \mathcal{U}$)
    \\ 
    $\sigma(\bx, \bm{c})$ & ($\dim \mathcal{X} + \dim \bm{c}$, 50,
                            $\dim \mathcal{U}$)
    \\ 
    $\bm{m}(\bx, \bm{c})$ & ($\dim \mathcal{X} + \dim \bm{c}$,
                            50, $\dim \mathcal{X}$)
    \\ 
    $V(\bx, \bm{c})$ & ($\dim \mathcal{X} + \dim \bm{c}$, 100, 1)
    \\ 
    $Q(\bx, \bu, \bm{c})$ & ($\dim \mathcal{X} + \dim \mathcal{U} + \dim \bm{c}$,
                            256, 1)
    \\
    $g(\bx, \bm{c})$ & ($\dim \mathcal{X} + \dim \bm{c}$, 50, 1)
    \\
    $D(\bx, \bm{c})$ & ($\dim \mathcal{X} + \dim \bm{c}$,
                       100, 1)
    \\
    $D(\bx, \bx', \bm{c})$ & ($2 \times \dim \mathcal{X} + \dim \bm{c}$,
                             256, 1)
    \\ \hline
  \end{tabular}
\end{table}

\subsection{Experimental results}
\label{sec:human_balancing_results}

Figure~\ref{fig:human:performance} shows the learning curves of the
seven subjects, indicating quite 
different learning processes. 
Subject 7 achieved the best performance for both conditions, and
subjects 1 and 3 failed to accomplish the task. Subjects 1 and 5 
performed well in the long-pole condition, but they failed to balance 
the pole in the upright position. 
Although the trajectories generated by Subjects 1 and 3 were 
unsuccessful, we used their data as the expert trajectories.

To evaluate the algorithms, we computed the NLL
for test datasets $\mathcal{D}_{i,\mathrm{te}}^E$:
\begin{displaymath}
  \mathrm{NLL}(i,j) = - \frac{1}{N_{i, \mathrm{te}}^E} 
  \sum_{\ell=1}^{N_{i, j, \mathrm{te}}^E} \ln \pi^L (\bm{x}_\ell' \mid \bm{x}_\ell,
  \bm{c}_{i,j}),
\end{displaymath}
where $N_{i, \mathrm{te}}^E$ is the number of samples in
$\mathcal{D}_{i, \mathrm{te}}^E$. $\bm{c}_{i,j} = [\bm{c}_i^s, \bm{c}_j^c]$
is the conditioning vector, where $\bm{c}_i$ represents that
the $i$-th element is 1 and otherwise 0.
See \ref{sec:linear_transformation} for an evaluation of the
state transition probability. 
Fig.~\ref{fig:human:nll_methods} shows that ERIL($\kappa^{-1} = 0$)
obtained a smaller NLL than the other baselines
for both conditions. Note that the data in the expert dataset were not
even sub-optimal because Subjects 3 and 5 did not fully accomplish the
balancing task (Fig.~\ref{fig:human:performance}).
The ERIL with IDM achieved comparable performance to ERIL ($\kappa^{-1} = 0$)
in the long-pole condition, but a lower performance in the
short-pole condition. This result suggests that data augmentation by IDM did not
help the discriminator’s training.
BCO also augmented the data by IDM, but its NLL
was higher than ERIL with IDM. Compared to the IRLGAN
and GAIfO results, our methods efficiently represented the policy of the experts,
 indicating that the structure of the ERIL's discriminator was critical
even if expert actions were unavailable.

Figure~\ref{fig:human:rewards} shows the reward function of 
Subjects 2, 4, and 7, estimated by ERIL($\kappa^{-1} = 0$), which was
projected to subspace $(\theta, \dot{\theta})$; $x, z, \dot{x}$
and $\dot{z}$ were set to zero for visualization. 
Although they balanced the pole in the long-pole condition, their estimated 
rewards differed. 
In the case of Subject 7, the reward function of the long-pole condition was 
about the same as the short-pole condition, although there was a 
significant difference in the results of Subject 4, who did not
perform well in the short-pole condition
(Fig.~\ref{fig:human:performance}).

Finally, we computed NLL where the policy
of the $i$-th subject was evaluated by the test dataset of the $j$-th
subject.
Fig.~\ref{fig:human:nll_subjects} shows the 
ERIL ($\kappa^{-1} = 0$), GAIfO, and BCO results. In the upper row,
we used the test dataset of Subject 4 in the long-pole condition.
For instance, the upper left figure shows the set of NLL
computed by
\begin{displaymath}
  \mathrm{NLL}(i,j) = - \frac{1}{N_{4, 1, \mathrm{te}}^E} 
  \sum_{\ell=1}^{N_{4, 1, \mathrm{te}}^E} \ln \pi^L
  (\bm{x}_\ell' \mid \bm{x}_\ell, \bm{c}_{i,j}).
\end{displaymath}
Note that the test dataset and the training data were inconsistent
when $i \neq 4$ and $j \neq 1$. The minimum NLL
was achieved when the conditioning vector was set properly
(i.e., $i = 4$ and $j = 1$). 
On the other hand, the policy of Subject 4 in the short-pole condition
($i = 4$ and $j = 2$) did not fit the test data well in the long-pole
condition, suggesting that Subject 4 used different reward functions
for different poles.

The bottom row in the figures shows the results when we used the test dataset
of Subject 7 in the long-pole condition. The minimum NLL
was achieved by a proper conditioning vector.
Unlike the case of Subject 4, ERIL ($\kappa^{-1} = 0$) found
that the policy for the short-pole condition ($i =7$ and $j = 2$)
also achieved better performance with no significant
differences. These results suggest that Subject 7, who was the best
performer, used similar reward functions for both poles.
No such property was found in GAIfO and BCO.

\section{Discussion}
\label{sec:discussion}

\subsection{Hyperparameter settings}

ERIL has two hyperparameters,
$\kappa$ and $\eta$.
As with a standard RL,
efficiency and performance depend on how the hyperparameters are tuned
during learning \citep{Henderson2018b, Zhang2021a}.
\citet{Kozuno2019a} showed that one hyperparameter derived from
$\kappa$ and $\eta$ controlled the tradeoff between the noise tolerance and the
convergence rate, and beta controls the quality of the asymptotic
performance from the viewpoint of the forward RL setting.
  We adopted a simple grid search method to tune the hyperparameters
  for every task, although naive grid search methods are sample
  inefficient and computationally expensive.

  One possible way to tune the hyperparameters is to evaluate
  multiple hyperparameters and update them using a genetic
  algorithm-like method \citep{Elfwing2018b, Jaderberg2017a}.
  A later version of SAC \citep{Haarnoja2018c} updates the
  hyperparameter that corresponds to the $\kappa$ of ERIL by a
  simple gradient descent algorithm.
  \citet{Lee2020a} optimized the state-dependent hyperparameter that
  corresponds to $\eta$ by the hypergradient on the
  validation data. Their methods will be helpful for the
  forward RL step.

However, the ERIL hyperparameters play a different role.
The $\kappa$ of ERIL, which is the coefficient of the entropy term of the
expert policy, should be determined by the properties of
the expert policy.
On the other hand, the $\kappa$ of entropy-regularized RL,
which is the coefficient
of the learner's policy, determines the softness of the max
operator of the Bellman optimality equation.
The $\eta$ of ERIL is the coefficient of the KL divergence between the
expert and the learner policies, and the
entropy-regularized RL is the coefficient of the KL divergence
between the learner's current policy and the previous one.
Tuning hyperparameters is future research. 

\subsection{Kinesthetic teaching}

ERIL assumes that experts and learners have identical state
transition probabilities. Therefore, the log-ratios of joint
distributions can be decomposed into the sum of the
policies and the state distributions.
However, this assumption limits ERIL's applicability.
For example, for a humanoid robot to imitate human
expert behavior, the expert must demonstrate by
kinesthetic teaching. To do so, the robot's gravity compensation mode is activated, and the expert must guide the robot's body to
execute the task.
Since this might not be a natural behavior for the expert, 
kinesthetic teaching is neither a practical nor a viable option to provide
demonstrations.

To overcome this problem, we should consider the log-ratio between
the state transitions of experts and learners. The log-ratio can be
estimated by the density ratio estimation methods, as we did for
the first discriminator.
We plan to modify ERIL to incorporate the third discriminator. 

\section{Conclusion}
\label{sec:conclusion}

This paper presented ERIL, which is entropy-regularized imitation
learning based on forward and inverse reinforcement learning.
Unlike previous methods, the update rules of the forward and
inverse RL steps are derived from the same soft Bellman equation,
and the state value function and hyperparameters are shared
between the inverse and forward RL steps.
The inverse RL step is done by training two binary classifiers,
one of which is constructed by the reward function, the state value
function, and the learner's policy.
Therefore, the state value function estimated by the inverse RL
step is used to initialize the state value function of the
forward RL step. The state value function updated by
the forward RL step also provides an initial state value function
of the inverse RL step.

Experimental results of the MuJoCo control benchmarks show that
ERIL was more sample-efficient than the modern off-policy imitation
learning algorithms in terms of environmental interactions
in the forward RL step.
  We also showed that ERIL got better asymptotic performance in
  a vision-based, target-reaching task whose experimental results
  demonstrated the importance of the first discriminator, which
  does not appear in other imitation learning methods.
ERIL also showed comparable performance
in terms of the number of demonstrations provided by the expert. 
In pole-balancing experiments, we showed how ERIL can be
applied to the analysis of human behaviors.

\appendix
\section{Derivation}
\label{sec:derivation}

\subsection{Derivation of Eq.~(\ref{eq:ERIL:Bellman2})}
\label{sec:eq:ERIL:Bellman2}

Since we assume that agent-environment interaction is modeled as a
Markov decision process, joint distribution is decomposed by
Eq.~(\ref{eq:ERIL:joint_prob_decomposition}). 
The log of the density ratio is given by
\begin{equation}
  \label{eq:APP:log_joint_decomposition}
  \ln \frac{\pi_k^L(\bx, \bu, \bx')}{\pi^E(\bx, \bu, \bx')} =
  \ln \frac{\pi_k^L(\bu \mid \bx)}{\pi^E(\bu \mid \bx)} +
  \ln \frac{\pi_k^L(\bx)}{\pi^E(\bx)}.
\end{equation}
Assign selector variable $L = 1$ to the samples from the
learner at the $k$-th iteration and $L = -1$ to the samples from
the expert:
\begin{align}
  \pi^L_k (\bx) &\triangleq \Pr (\bx \mid L = 1), \notag \\
  \pi^E (\bx) &\triangleq \Pr (\bx \mid L = -1) = 1 - \pi_k^L(\bx). \notag 
\end{align}
Then the first discriminator is represented by
\begin{displaymath}
  D_k^{(1)}(\bx) \triangleq \frac{\Pr (L = 1 \mid \bx) \Pr (L = 1)}
  {\Pr (\bx)}.
\end{displaymath}
We obtain the following log of the density ratio:
\begin{equation}
  \label{eq:APP:density_ratio}
  \ln \frac{\pi_k^L(\bx)}{\pi^E(\bx)} = 
  \ln \frac{D_k^{(1)}(\bx)}{1 - D_k^{(1)}(\bx)}
  - 
  \ln \frac{\Pr (L = 1)}{\Pr (L = -1)}.
\end{equation}
The log of density ratio
$\ln \pi_k^L(\bx, \bu, \bx') / \pi^L(\bx, \bu, \bx')$ is represented by
second discriminator $D_k^{(2)}(\bx, \bu, \bx')$ in the same way.
Substituting the above results and Eq.~(\ref{eq:ERIL:Bellman1})
into Eq.~(\ref{eq:APP:log_joint_decomposition})
yields Eq.~(\ref{eq:ERIL:Bellman2}). 

\subsection{Derivation of Eq.~(\ref{eq:ERIL:optimal_discriminator})}
\label{sec:eq:ERIL:optimal_discriminator}

The probability ratio of the learner and expert samples is simply
estimated by the ratio of the number of samples \citep{Sugiyama2012b}:
\begin{displaymath}
  \ln \frac{\Pr (L = 1)}{\Pr (L = -1)} \approx
  \ln \frac{\lvert \mathcal{D}^L \rvert}{\lvert \mathcal{D}^E \rvert}.
\end{displaymath}
When $\lvert \mathcal{D}^L \rvert = \lvert \mathcal{D}^E \rvert$,
$\exp (g_k(\bx))$ represents the following density ratio:
\begin{displaymath}
  \exp (g_k(\bx)) = \pi_k^L(\bx) / \pi^E(\bx). 
\end{displaymath}
Arranging Eq.~(\ref{eq:ERIL:Bellman3}) yields
\begin{displaymath}
  D^{(2)}(\bx, \bu) =
  \frac{\pi_k^L(\bu \mid \bx)}
  {\exp (-g_k(\bx)) \tilde{\pi}^E (\bu \mid \bx) + \pi_k^L (\bu \mid \bx)},
\end{displaymath}
and we immediately obtain Eq.~(\ref{eq:ERIL:optimal_discriminator}).
When term $\exp (-g_k(\bx))$ is ignored,
$D^{(2)}(\bx, \bu)$ represents the optimal discriminator conditioned
on the state.

\section{Deterministic Regularized Autoencoders}
\label{sec:RAE}

  This appendix briefly explains the deterministic RAE \citep{Ghosh2020a}.
  For notational brevity, captured images and latent variables
  are denoted by $\bx$ and $\bm{z}$.
  Suppose that the encoder network is given by mean
  $\bm{\mu}_e$ and covariance parameters $\bm{\sigma}_e$:
  \begin{displaymath}
    E(\bx) = \bm{\mu}_e(\bx) + \bm{\sigma}_e(\bx)
    \odot \bm{\epsilon}, \quad
    \bm{\epsilon} \sim \mathcal{N}(\bm{0}, \bm{I}),
  \end{displaymath}
  where $\odot$ denotes the Hadamard product.
  $\bm{I}$ is the identity matrix.
  The decoder network is also given by $\bm{\mu}_d$ and
  $\bm{\sigma}_d$ in the same way.
  The loss function of RAE is given by
  \begin{displaymath}
    J_{\mathrm{RAE}}(\bw_e, \bw_d) =
    \mathbb{E}_{\bx \sim \mathcal{D}} \left[
      \| \bx - \bm{\mu}_d (E(\bx)) \|_2^2 + \lambda_2
      \| \bm{z} \|_2^2 + \lambda_d \| \bw_d \|_2^2
    \right],
  \end{displaymath}
  where $\lambda_e$ and $\lambda_d$ are hyperparameters.
  We set $\lambda_e = 10^{-6}$ and $\lambda_d = 10^{-7}$
  according to a previous work \citep{Yarats2020a}. Variables $\bw_e$ and
  $\bw_d$ represent the network weights of the encoder and
  the decoder.

\section{Notes on the Pole-balancing Problem}

\subsection{Equations of motion}
\label{sec:xz_inverted_pendulum}

Figure~\ref{fig:human:task}(b) shows the X-Z inverted pendulum on a pivot
driven by horizontal and vertical forces. We used the equations of motion
given in a previous work \citep{Wang2012a}:
\begin{align}
  M(M + m) \ddot{x} &= Mml \dot{\theta}^2 \sin \theta +
    (M + m \cos^2 \theta) F_x - m F_z \sin \theta \cos \theta,
  \notag \\
  \label{eq:pendulum:equation}
  M(M + m) \ddot{z} &= Mml \dot{\theta}^2 \cos \theta -
    (M + m \sin^2 \theta) F_z - g,
  \\
  Ml \ddot{\theta} &= - F_x \cos \theta + F_z \sin \theta,
  \notag 
\end{align}
where $(x, z)$ is the position of the pivot in the $xoz$ coordinate
and $(\dot{x}, \dot{z})$ and $(\ddot{x}, \ddot{z})$ are the speed
and acceleration.
$M$ and $m$ are the mass of the pivot and the pendulum.
$l$ is the distance from the pivot to the center of the
pendulum's mass. $g$ denotes the gravitational acceleration.
$F_x$ and $F_z$ are the horizontal and vertical forces.

The state vector is given by a three-dimensional vector,
$\bm{x} = [x, \dot{x}, z, \dot{z}, \theta, \dot{\theta}]^\top$, and
the action is represented by a two-dimensional vector,
$\bm{u} = [F_x, F_z]^\top$.
To implement the simulation, the time axis is discretized by $h = 0.01$ [s]. 
The parameters of the X-Z inverted pendulum are described below:
$M = 0.85$ [kg], $m = 0.30$ [kg] (long pole) or $0.12$ [kg] (short pole),
and $l = 0.73$ [m] (long pole) or $0.29$ [m] (short pole).
$g = 9.81$ [$\mathrm{m/s^2}$]. 
The inertia of the pendulum is negligible.

\subsection{Evaluation of state transition probability}
\label{sec:linear_transformation}

Since no action is available in the pole-balancing task,
the learner's policy cannot be used for evaluation. We exploit the
property where a linear transformation of a multivariate Gaussian
random vector has a multivariate Gaussian distribution because
the policy in this study is also represented by a Gaussian
distribution. Using a first-order Taylor expansion, the
time-discretized version of Eq.~(\ref{eq:pendulum:equation}) can
be expressed by $\bm{x}_{t+1} = \bm{A}_t \bm{x}_t + \bm{B}_t \bm{u}_t$. 

\subsection*{Acknowledgments}

This work was supported by Innovative Science and Technology Initiative
for Security Grant Number JPJ004596, ATLA, Japan and
JST-Mirai Program Grant Number JPMJMI18B8, Japan to EU.
This work was also supported by the Japan Society for the Promotion of
Science, KAKENHI Grant Numbers JP16K21738, JP16H06561, and JP16H06563
to KD and JP19H05001 to EU.
We thank Shoko Igarashi who collected the data of the
human pole-balancing task. 

\section*{Figure captions}

Figure~\ref{fig:eril_interaction}:
Overall architecture of Entropy-Regularized Imitation Learning (ERIL)

Figure~\ref{fig:eril_discriminator}:
Architecture of two ERIL discriminators: $D_k^{(1)}(\bx)$ and
$D_k^{(2)}(\bx, \bu, \bx')$.

Figure~\ref{fig:mujoco:frl_normal}:
Performance with respect to number of interactions
in MuJoCo tasks: Solid lines
represent average values, and shaded areas correspond to $\pm 1$
standard deviation region.
    
Figure~\ref{fig:mujoco:irl_normal}:
Performance with respect to number of trajectories provided by expert
in MuJoCo tasks:
Solid lines represent average values, and shaded areas
correspond to $\pm 1$ standard deviation region.

Figure~\ref{fig:mujoco:frl_ablation}:
Comparison of learning curves in ablation study

  Figure~\ref{fig:nextage:task}:
  Real robot setting: (a) environment of reaching task, which
  moves the end-effector of left arm to 
  target position; (b) possible environmental configuration.

  Figure~\ref{fig:nextage:network}:
  Neural networks used in robot experiment:
  (a) architecture for policy, reward, state value,
  state-action value, and first discriminator, sharing
  the encoder;
  (b) architecture for encoder: Conv. denotes a
  convolutional neural network. ``$n$c'' denotes
 ``$n$ channels.''

  Figure~\ref{fig:nextage:performance}:
  Performance with respect to number of interactions
  in real robot experiment 

  Figure~\ref{fig:nextage:nll}:
  Comparison of NLL in real robot experiment:
  ``ERIL w/o D(1)'' denotes ERIL without first discriminator.
  Note that smaller NLL values indicate a better fit.

Figure~\ref{fig:human:task}:
Inverted pendulum task solved by a human subject: (a) start and
goal positions; (b) state representation.
Notations are explained in
  \ref{sec:xz_inverted_pendulum}.

Figure~\ref{fig:human:performance}:
Learning curves of seven subjects: blue: long pole; red: short pole.
Trial is considered a failure if subject cannot balance pole in 
upright position within 40 [s].

Figure~\ref{fig:human:nll_methods}:
Comparison of NLL among imitation learning
algorithms: Note that smaller NLL values indicate a better fit.
  
Figure~\ref{fig:human:rewards}:
Estimated reward function of Subjects 2, 4, and 7 projected to subspace
$(\theta, \omega)$, $\omega = d\theta/dt$

Figure~\ref{fig:human:nll_subjects}: 
Comparison of NLL when condition of test
dataset was different from training one 
in human inverted pendulum task. 
Figs. in upper row (a, b, and c) show results when 
test dataset is $\mathcal{D}_{4, 1, \mathrm{te}}^E$.
Figs. in lower row (d, e, and f) show results when 
test dataset is $\mathcal{D}_{7, 1, \mathrm{te}}^E$.

\vskip 0.2in
\bibliographystyle{elsarticle-num-names}
\bibliography{./reference.bib}

\begin{thebibliography}{79}
\expandafter\ifx\csname natexlab\endcsname\relax\def\natexlab#1{#1}\fi
\providecommand{\url}[1]{\texttt{#1}}
\providecommand{\href}[2]{#2}
\providecommand{\path}[1]{#1}
\providecommand{\DOIprefix}{doi:}
\providecommand{\ArXivprefix}{arXiv:}
\providecommand{\URLprefix}{URL: }
\providecommand{\Pubmedprefix}{pmid:}
\providecommand{\doi}[1]{\href{http://dx.doi.org/#1}{\path{#1}}}
\providecommand{\Pubmed}[1]{\href{pmid:#1}{\path{#1}}}
\providecommand{\bibinfo}[2]{#2}
\ifx\xfnm\relax \def\xfnm[#1]{\unskip,\space#1}\fi
\bibitem[{Sutton and Barto(1998)}]{Sutton1998a}
\bibinfo{author}{R.~Sutton}, \bibinfo{author}{A.~Barto},
  \bibinfo{title}{{Reinforcement Learning}}, \bibinfo{publisher}{MIT Press},
  \bibinfo{year}{1998}.
\bibitem[{Doya(2007)}]{Doya2007a}
\bibinfo{author}{K.~Doya},
\newblock \bibinfo{title}{Reinforcement learning: Computational theory and
  biological mechanisms},
\newblock \bibinfo{journal}{HFSP Journal} \bibinfo{volume}{1}
  (\bibinfo{year}{2007}) \bibinfo{pages}{30--40}.
\bibitem[{Kober et~al.(2013)Kober, Bagnell, and Peters}]{Kober2013a}
\bibinfo{author}{J.~Kober}, \bibinfo{author}{J.~A. Bagnell},
  \bibinfo{author}{J.~Peters},
\newblock \bibinfo{title}{{Reinforcement Learning in Robotics: A Survey}},
\newblock \bibinfo{journal}{International Journal of Robotics Research}
  \bibinfo{volume}{32} (\bibinfo{year}{2013}) \bibinfo{pages}{1238--1274}.
\bibitem[{Mnih et~al.(2015)Mnih, Kavukcuoglu, Silver, Rusu, Veness, Bellemare,
  Graves, Riedmiller, Fidjeland, Ostrovski et~al.}]{Mnih2015a}
\bibinfo{author}{V.~Mnih}, \bibinfo{author}{K.~Kavukcuoglu},
  \bibinfo{author}{D.~Silver}, \bibinfo{author}{A.~A. Rusu},
  \bibinfo{author}{J.~Veness}, \bibinfo{author}{M.~G. Bellemare},
  \bibinfo{author}{A.~Graves}, \bibinfo{author}{M.~Riedmiller},
  \bibinfo{author}{A.~K. Fidjeland}, \bibinfo{author}{G.~Ostrovski}, et~al.,
\newblock \bibinfo{title}{Human-level control through deep reinforcement
  learning},
\newblock \bibinfo{journal}{Nature} \bibinfo{volume}{518}
  (\bibinfo{year}{2015}) \bibinfo{pages}{529--533}.
\bibitem[{Silver et~al.(2017)Silver, Schrittwieser, Simonyan, Antonoglou,
  Huang, Guez, Hubert et~al.}]{Silver2017a}
\bibinfo{author}{D.~Silver}, \bibinfo{author}{J.~Schrittwieser},
  \bibinfo{author}{K.~Simonyan}, \bibinfo{author}{I.~Antonoglou},
  \bibinfo{author}{A.~Huang}, \bibinfo{author}{A.~Guez},
  \bibinfo{author}{T.~Hubert}, et~al.,
\newblock \bibinfo{title}{Mastering the game of go without human knowledge},
\newblock \bibinfo{journal}{Nature} \bibinfo{volume}{550}
  (\bibinfo{year}{2017}) \bibinfo{pages}{354--59}.
\bibitem[{OpenAI et~al.(2019)OpenAI, Berner, Brockman, Chan, Cheung, Debiak,
  Dennison et~al.}]{OpenAI2019a}
\bibinfo{author}{OpenAI}, \bibinfo{author}{C.~Berner},
  \bibinfo{author}{G.~Brockman}, \bibinfo{author}{B.~Chan},
  \bibinfo{author}{V.~Cheung}, \bibinfo{author}{P.~Debiak},
  \bibinfo{author}{C.~Dennison}, et~al.,
\newblock \bibinfo{title}{Dota 2 with large scale deep reinforcement learning},
\newblock \bibinfo{journal}{arXiv preprint}  (\bibinfo{year}{2019}).
\bibitem[{Vinyals et~al.(2019)Vinyals, Babuschkin, Czarnecki, Mathieu, Dudzik,
  Chung, Choi et~al.}]{Vinyals2019a}
\bibinfo{author}{O.~Vinyals}, \bibinfo{author}{I.~Babuschkin},
  \bibinfo{author}{W.~M. Czarnecki}, \bibinfo{author}{M.~Mathieu},
  \bibinfo{author}{A.~Dudzik}, \bibinfo{author}{J.~Chung},
  \bibinfo{author}{D.~H. Choi}, et~al.,
\newblock \bibinfo{title}{Grandmaster level in {StarCraft II} using multi-agent
  reinforcement learning},
\newblock \bibinfo{journal}{Nature} \bibinfo{volume}{575}
  (\bibinfo{year}{2019}) \bibinfo{pages}{350--54}.
\bibitem[{OpenAI et~al.(2019)OpenAI, Akkaya, Andrychowicz, Chociej, Litwin,
  McGrew, Petron et~al.}]{OpenAI2019b}
\bibinfo{author}{OpenAI}, \bibinfo{author}{I.~Akkaya},
  \bibinfo{author}{M.~Andrychowicz}, \bibinfo{author}{M.~Chociej},
  \bibinfo{author}{M.~Litwin}, \bibinfo{author}{B.~McGrew},
  \bibinfo{author}{A.~Petron}, et~al.,
\newblock \bibinfo{title}{Solving rubik’s cube with a robot hand},
\newblock \bibinfo{journal}{arXiv preprint}  (\bibinfo{year}{2019}).
\bibitem[{Tsurumine et~al.(2019)Tsurumine, Cui, Uchibe, and
  Matsubara}]{Tsurumine2019a}
\bibinfo{author}{Y.~Tsurumine}, \bibinfo{author}{Y.~Cui},
  \bibinfo{author}{E.~Uchibe}, \bibinfo{author}{T.~Matsubara},
\newblock \bibinfo{title}{Deep reinforcement learning with smooth policy
  update: Application to robotic cloth manipulation},
\newblock \bibinfo{journal}{Robotics and Autonomous Systems}
  \bibinfo{volume}{112} (\bibinfo{year}{2019}) \bibinfo{pages}{72--83}.
\bibitem[{Haarnoja et~al.(2018)Haarnoja, Zhou, Hartikainen, Tucker, Ha, Tan,
  Kumar, Zhu, Gupta, Abbeel, and Levine}]{Haarnoja2018c}
\bibinfo{author}{T.~Haarnoja}, \bibinfo{author}{A.~Zhou},
  \bibinfo{author}{K.~Hartikainen}, \bibinfo{author}{G.~Tucker},
  \bibinfo{author}{S.~Ha}, \bibinfo{author}{J.~Tan},
  \bibinfo{author}{V.~Kumar}, \bibinfo{author}{H.~Zhu},
  \bibinfo{author}{A.~Gupta}, \bibinfo{author}{P.~Abbeel},
  \bibinfo{author}{S.~Levine},
\newblock \bibinfo{title}{Soft actor-critic algorithms and applications},
\newblock \bibinfo{journal}{arXiv preprint}  (\bibinfo{year}{2018}).
\bibitem[{Wang and Qiao(2019)}]{Wang2019a}
\bibinfo{author}{D.~Wang}, \bibinfo{author}{J.~Qiao},
\newblock \bibinfo{title}{Approximate neural optimal control with reinforcement
  learning for a torsional pendulum device},
\newblock \bibinfo{journal}{Neural Networks} \bibinfo{volume}{117}
  (\bibinfo{year}{2019}) \bibinfo{pages}{1--7}.
\bibitem[{Odekunle et~al.(2020)Odekunle, W.~Gao, and Jiang}]{Odekunle2020a}
\bibinfo{author}{A.~Odekunle}, \bibinfo{author}{M.~D. W.~Gao},
  \bibinfo{author}{Z.-P. Jiang},
\newblock \bibinfo{title}{Reinforcement learning and non-zero-sum game output
  regulation for multi-player linear uncertain systems},
\newblock \bibinfo{journal}{Automatica} \bibinfo{volume}{112}
  (\bibinfo{year}{2020}).
\bibitem[{Doya and Uchibe(2005)}]{Doya2005a}
\bibinfo{author}{K.~Doya}, \bibinfo{author}{E.~Uchibe},
\newblock \bibinfo{title}{{The Cyber Rodent} project: Exploration of adaptive
  mechanisms for self-preservation and self-reproduction},
\newblock \bibinfo{journal}{Adaptive Behavior} \bibinfo{volume}{13}
  (\bibinfo{year}{2005}) \bibinfo{pages}{149--60}.
\bibitem[{Ross et~al.(2011)Ross, Gordon, and Bagnell}]{Ross2011a}
\bibinfo{author}{S.~Ross}, \bibinfo{author}{G.~Gordon},
  \bibinfo{author}{D.~Bagnell},
\newblock \bibinfo{title}{A reduction of imitation learning and structured
  prediction to no-regret online learning},
\newblock in: \bibinfo{booktitle}{Proc. of the 14th International Conference on
  Artificial Intelligence and Statistics}, \bibinfo{year}{2011}, pp.
  \bibinfo{pages}{627--35}.
\bibitem[{Ng and Russell(2000)}]{Ng2000b}
\bibinfo{author}{A.~Y. Ng}, \bibinfo{author}{S.~Russell},
\newblock \bibinfo{title}{Algorithms for inverse reinforcement learning},
\newblock in: \bibinfo{booktitle}{Proc. of the 17th International Conference on
  Machine Learning}, \bibinfo{year}{2000}.
\bibitem[{Abbeel and Ng(2004)}]{Abbeel2004a}
\bibinfo{author}{P.~Abbeel}, \bibinfo{author}{A.~Y. Ng},
\newblock \bibinfo{title}{Apprenticeship learning via inverse reinforcement
  learning},
\newblock in: \bibinfo{booktitle}{Proc. of the 21st International Conference on
  Machine Learning}, \bibinfo{year}{2004}.
\bibitem[{Vogel et~al.(2012)Vogel, Ramachandran, Gupta, and Raux}]{Vogel2012a}
\bibinfo{author}{A.~Vogel}, \bibinfo{author}{D.~Ramachandran},
  \bibinfo{author}{R.~Gupta}, \bibinfo{author}{A.~Raux},
\newblock \bibinfo{title}{{Improving Hybrid Vehicle Fuel Efficiency using
  Inverse Reinforcement Learning}},
\newblock in: \bibinfo{booktitle}{Proc. of the 26th AAAI Conference on
  Artificial Intelligence}, \bibinfo{year}{2012}.
\bibitem[{Liu et~al.(2013)Liu, Araujo, Brunskill, Rossetti, Barros, and
  Krishnan}]{Liu2013b}
\bibinfo{author}{S.~Liu}, \bibinfo{author}{M.~Araujo},
  \bibinfo{author}{E.~Brunskill}, \bibinfo{author}{R.~Rossetti},
  \bibinfo{author}{J.~Barros}, \bibinfo{author}{R.~Krishnan},
\newblock \bibinfo{title}{Understanding sequential decisions via inverse
  reinforcement learning},
\newblock in: \bibinfo{booktitle}{Proc. of the 14th IEEE International
  Conference on Mobile Data Management}, \bibinfo{publisher}{IEEE},
  \bibinfo{year}{2013}, pp. \bibinfo{pages}{177--186}.
\bibitem[{Shimosaka et~al.(2014)Shimosaka, Kaneko, and Nishi}]{Shimosaka2014a}
\bibinfo{author}{M.~Shimosaka}, \bibinfo{author}{T.~Kaneko},
  \bibinfo{author}{K.~Nishi},
\newblock \bibinfo{title}{{Modeling risk anticipation and defensive driving on
  residential roads with inverse reinforcement learning}},
\newblock in: \bibinfo{booktitle}{Proc. of the 17th International IEEE
  Conference on Intelligent Transportation Systems}, \bibinfo{year}{2014}, pp.
  \bibinfo{pages}{1694--1700}.
\bibitem[{Muelling et~al.(2014)Muelling, Boularias, Mohler, Sch\"{o}lkopf, and
  Peters}]{Muelling2014a}
\bibinfo{author}{K.~Muelling}, \bibinfo{author}{A.~Boularias},
  \bibinfo{author}{B.~Mohler}, \bibinfo{author}{B.~Sch\"{o}lkopf},
  \bibinfo{author}{J.~Peters},
\newblock \bibinfo{title}{{Learning strategies in table tennis using inverse
  reinforcement learning.}},
\newblock \bibinfo{journal}{Biological Cybernetics} \bibinfo{volume}{108}
  (\bibinfo{year}{2014}) \bibinfo{pages}{603--619}.
\bibitem[{Kretzschmar et~al.(2016)Kretzschmar, Spies, Sprunk, and
  Burgard}]{Kretzschmar2016a}
\bibinfo{author}{H.~Kretzschmar}, \bibinfo{author}{M.~Spies},
  \bibinfo{author}{C.~Sprunk}, \bibinfo{author}{W.~Burgard},
\newblock \bibinfo{title}{{Socially compliant mobile robot navigation via
  inverse reinforcement learning}},
\newblock \bibinfo{journal}{The International Journal of Robotics Research}
  (\bibinfo{year}{2016}).
\bibitem[{Xia and {El Kamel}(2016)}]{Xia2016a}
\bibinfo{author}{C.~Xia}, \bibinfo{author}{A.~{El Kamel}},
\newblock \bibinfo{title}{Neural inverse reinforcement learning in autonomous
  navigation},
\newblock \bibinfo{journal}{Robotics and Autonomous Systems}
  \bibinfo{volume}{84} (\bibinfo{year}{2016}) \bibinfo{pages}{1--14}.
\bibitem[{Ashida et~al.(2019)Ashida, Kato, Hotta, and Oka}]{Ashida2019a}
\bibinfo{author}{K.~Ashida}, \bibinfo{author}{T.~Kato},
  \bibinfo{author}{K.~Hotta}, \bibinfo{author}{K.~Oka},
\newblock \bibinfo{title}{Multiple tracking and machine learning reveal
  dopamine modulation for area-restricted foraging behaviors via velocity
  change in caenorhabditis elegans},
\newblock \bibinfo{journal}{Neuroscience Letters} \bibinfo{volume}{706}
  (\bibinfo{year}{2019}) \bibinfo{pages}{68--74}.
\bibitem[{Hirakawa et~al.(2018)Hirakawa, Yamashita, Tamaki, Fujiyoshi, Umezu,
  Takeuchi, Matsumoto, and Yoda}]{Hirakawa2018a}
\bibinfo{author}{T.~Hirakawa}, \bibinfo{author}{T.~Yamashita},
  \bibinfo{author}{T.~Tamaki}, \bibinfo{author}{H.~Fujiyoshi},
  \bibinfo{author}{Y.~Umezu}, \bibinfo{author}{I.~Takeuchi},
  \bibinfo{author}{S.~Matsumoto}, \bibinfo{author}{K.~Yoda},
\newblock \bibinfo{title}{Can ai predict animal movements? filling gaps in
  animal trajectories using inverse reinforcement learning},
\newblock \bibinfo{journal}{Ecosphere}  (\bibinfo{year}{2018}).
\bibitem[{Yamaguchi et~al.(2018)Yamaguchi, Honda, Ikeda, Nakano, Mori, and
  Ishii}]{Yamaguchi2018a}
\bibinfo{author}{S.~Yamaguchi}, \bibinfo{author}{N.~Honda},
  \bibinfo{author}{Y.~Ikeda, M.~Tsukada}, \bibinfo{author}{S.~Nakano},
  \bibinfo{author}{I.~Mori}, \bibinfo{author}{S.~Ishii},
\newblock \bibinfo{title}{Identification of animal behavioral strategies by
  inverse reinforcement learning},
\newblock \bibinfo{journal}{PLoS Computational Biology}
  (\bibinfo{year}{2018}).
\bibitem[{Neu and Szepesv\'{a}ri(2009)}]{Neu2009a}
\bibinfo{author}{G.~Neu}, \bibinfo{author}{C.~Szepesv\'{a}ri},
\newblock \bibinfo{title}{{Training parsers by inverse reinforcement
  learning}},
\newblock \bibinfo{journal}{Machine Learning} \bibinfo{volume}{77}
  (\bibinfo{year}{2009}) \bibinfo{pages}{303--337}.
\bibitem[{Collette et~al.(2017)Collette, Pauli, Bossaerts, and
  O'Doherty}]{Collette2017a}
\bibinfo{author}{S.~Collette}, \bibinfo{author}{W.~M. Pauli},
  \bibinfo{author}{P.~Bossaerts}, \bibinfo{author}{J.~O'Doherty},
\newblock \bibinfo{title}{Neural computations underlying inverse reinforcement
  learning in the human brain},
\newblock \bibinfo{journal}{eLife} \bibinfo{volume}{6} (\bibinfo{year}{2017}).
\bibitem[{Fu et~al.(2018)Fu, Luo, and Levine}]{Fu2018a}
\bibinfo{author}{J.~Fu}, \bibinfo{author}{K.~Luo}, \bibinfo{author}{S.~Levine},
\newblock \bibinfo{title}{Learning robust rewards with adversarial inverse
  reinforcement learning},
\newblock in: \bibinfo{booktitle}{Proc. of the 6th International Conference on
  Learning Representations}, \bibinfo{year}{2018}.
\bibitem[{Ho and Ermon(2016)}]{Ho2016c}
\bibinfo{author}{J.~Ho}, \bibinfo{author}{S.~Ermon},
\newblock \bibinfo{title}{Generative adversarial imitation learning},
\newblock in: \bibinfo{booktitle}{Advances in Neural Information Processing
  Systems 29}, \bibinfo{year}{2016}, pp. \bibinfo{pages}{4565--73}.
\bibitem[{Goodfellow et~al.(2014)Goodfellow, Pouget-Abadie, Mirza, Xu,
  Warde-Farley, Ozair, Courville, and Bengio}]{Goodfellow2014a}
\bibinfo{author}{I.~Goodfellow}, \bibinfo{author}{J.~Pouget-Abadie},
  \bibinfo{author}{M.~Mirza}, \bibinfo{author}{B.~Xu},
  \bibinfo{author}{D.~Warde-Farley}, \bibinfo{author}{S.~Ozair},
  \bibinfo{author}{A.~Courville}, \bibinfo{author}{Y.~Bengio},
\newblock \bibinfo{title}{Generative adversarial nets},
\newblock in: \bibinfo{booktitle}{Advances in Neural Information Processing
  Systems 27}, \bibinfo{year}{2014}, pp. \bibinfo{pages}{2672--2680}.
\bibitem[{Jena et~al.(2020)Jena, Liu, and Sycara}]{Jena2020a}
\bibinfo{author}{R.~Jena}, \bibinfo{author}{C.~Liu},
  \bibinfo{author}{K.~Sycara},
\newblock \bibinfo{title}{Augmenting gail with bc for sample efficient
  imitation learning},
\newblock in: \bibinfo{booktitle}{Proc. of the 3rd Conference on Robot
  Learning}, \bibinfo{year}{2020}.
\bibitem[{Kinose and Taniguchi(2020)}]{Kinose2020a}
\bibinfo{author}{A.~Kinose}, \bibinfo{author}{T.~Taniguchi},
\newblock \bibinfo{title}{Integration of imitation learning using gail and
  reinforcement learning using task-achievement rewards via probabilistic
  graphical model},
\newblock \bibinfo{journal}{Advanced Robotics}  (\bibinfo{year}{2020})
  \bibinfo{pages}{1055--1067}.
\bibitem[{Sugiyama et~al.(2012)Sugiyama, Suzuki, and Kanamori}]{Sugiyama2012b}
\bibinfo{author}{M.~Sugiyama}, \bibinfo{author}{T.~Suzuki},
  \bibinfo{author}{T.~Kanamori}, \bibinfo{title}{Density ratio estimation in
  machine learning}, \bibinfo{publisher}{Cambridge University Press},
  \bibinfo{year}{2012}.
\bibitem[{Uchibe and Doya(2014)}]{Uchibe2014b}
\bibinfo{author}{E.~Uchibe}, \bibinfo{author}{K.~Doya},
\newblock \bibinfo{title}{{Inverse Reinforcement Learning Using Dynamic Policy
  Programming}},
\newblock in: \bibinfo{booktitle}{Proc. of IEEE International Conference on
  Development and Learning and Epigenetic Robotics}, \bibinfo{year}{2014}, pp.
  \bibinfo{pages}{222--228}.
\bibitem[{Uchibe(2018)}]{Uchibe2018c}
\bibinfo{author}{E.~Uchibe},
\newblock \bibinfo{title}{Model-free deep inverse reinforcement learning by
  logistic regression},
\newblock \bibinfo{journal}{Neural Processing Letters} \bibinfo{volume}{47}
  (\bibinfo{year}{2018}) \bibinfo{pages}{891--905}.
\bibitem[{Azar et~al.(2012)Azar, G{\'{o}}mez, and Kappen}]{Azar2012a}
\bibinfo{author}{M.~G. Azar}, \bibinfo{author}{V.~G{\'{o}}mez},
  \bibinfo{author}{H.~J. Kappen},
\newblock \bibinfo{title}{Dynamic policy programming},
\newblock \bibinfo{journal}{Journal of Machine Learning Research}
  \bibinfo{volume}{13} (\bibinfo{year}{2012}) \bibinfo{pages}{3207--3245}.
\bibitem[{Haarnoja et~al.(2018)Haarnoja, Zhou, Abbeel, and
  Levine}]{Haarnoja2018a}
\bibinfo{author}{T.~Haarnoja}, \bibinfo{author}{A.~Zhou},
  \bibinfo{author}{P.~Abbeel}, \bibinfo{author}{S.~Levine},
\newblock \bibinfo{title}{{Soft Actor-Critic: Off-policy maximum entropy deep
  reinforcement learning with a stochastic actor}},
\newblock in: \bibinfo{booktitle}{Proc. of the 35th International Conference on
  Machine Learning}, \bibinfo{year}{2018}, pp. \bibinfo{pages}{1856--1865}.
\bibitem[{Kozuno et~al.(2019)Kozuno, Uchibe, and Doya}]{Kozuno2019a}
\bibinfo{author}{T.~Kozuno}, \bibinfo{author}{E.~Uchibe},
  \bibinfo{author}{K.~Doya},
\newblock \bibinfo{title}{Theoretical analysis of efficiency and robustness of
  softmax and gap-increasing operators in reinforcement learning},
\newblock in: \bibinfo{booktitle}{Proc. of the 22nd International Conference on
  Artificial Intelligence and Statistics}, \bibinfo{address}{Okinawa, Japan},
  \bibinfo{year}{2019}, pp. \bibinfo{pages}{2995--3003}.
\bibitem[{Todorov et~al.(2012)Todorov, Erez, and Tassa}]{Todorov2012a}
\bibinfo{author}{E.~Todorov}, \bibinfo{author}{T.~Erez},
  \bibinfo{author}{Y.~Tassa},
\newblock \bibinfo{title}{{MuJoCo}: A physics engine for model-based control},
\newblock in: \bibinfo{booktitle}{Proc. of IEEE/RSJ International Conference on
  Intelligent Robots and Systems}, \bibinfo{year}{2012}, pp.
  \bibinfo{pages}{5026--5033}.
\bibitem[{Brockman et~al.(2016)Brockman, Cheung, Pettersson, Schneider,
  Schulman, Tang, and Zaremba}]{Brockman2016a}
\bibinfo{author}{G.~Brockman}, \bibinfo{author}{V.~Cheung},
  \bibinfo{author}{L.~Pettersson}, \bibinfo{author}{J.~Schneider},
  \bibinfo{author}{J.~Schulman}, \bibinfo{author}{J.~Tang},
  \bibinfo{author}{W.~Zaremba},
\newblock \bibinfo{title}{{OpenAI Gym}},
\newblock \bibinfo{journal}{arXiv preprint}  (\bibinfo{year}{2016}).
\bibitem[{Liu et~al.(2021)Liu, Li, Kang, and Darrell}]{Liu2021a}
\bibinfo{author}{Z.~Liu}, \bibinfo{author}{X.~Li}, \bibinfo{author}{B.~Kang},
  \bibinfo{author}{T.~Darrell},
\newblock \bibinfo{title}{Regularization matters in policy optimization -- an
  empirical study on continuous control},
\newblock in: \bibinfo{booktitle}{Proc. of the 9th International Conference on
  Learning Representations}, \bibinfo{year}{2021}.
\bibitem[{Parisi et~al.(2019)Parisi, Tangkaratt, Peters, and
  Khan}]{Parisi2019a}
\bibinfo{author}{S.~Parisi}, \bibinfo{author}{V.~Tangkaratt},
  \bibinfo{author}{J.~Peters}, \bibinfo{author}{M.~E. Khan},
\newblock \bibinfo{title}{{TD}-regularized actor-critic methods},
\newblock \bibinfo{journal}{Machine Learning}  (\bibinfo{year}{2019})
  \bibinfo{pages}{1467--1501}.
\bibitem[{Ohnishi et~al.(2019)Ohnishi, Uchibe, Yamaguchi, Nakanishi, Yasui, and
  Ishii}]{Ohnishi2019a}
\bibinfo{author}{S.~Ohnishi}, \bibinfo{author}{E.~Uchibe},
  \bibinfo{author}{Y.~Yamaguchi}, \bibinfo{author}{K.~Nakanishi},
  \bibinfo{author}{Y.~Yasui}, \bibinfo{author}{S.~Ishii},
\newblock \bibinfo{title}{Constrained deep {Q}-learning gradually approaching
  ordinary {Q}-learning},
\newblock \bibinfo{journal}{Frontiers in Neurorobotics} \bibinfo{volume}{13}
  (\bibinfo{year}{2019}).
\bibitem[{Li et~al.(2018)Li, Liu, and Wang}]{Li2018a}
\bibinfo{author}{H.~Li}, \bibinfo{author}{D.~Liu}, \bibinfo{author}{D.~Wang},
\newblock \bibinfo{title}{Manifold regularized reinforcement learning},
\newblock \bibinfo{journal}{IEEE Transactions on Neural Networks and Learning
  Systems} \bibinfo{volume}{29} (\bibinfo{year}{2018})
  \bibinfo{pages}{932--43}.
\bibitem[{Amit et~al.(2020)Amit, Meir, and Ciosek}]{Amit2020a}
\bibinfo{author}{R.~Amit}, \bibinfo{author}{R.~Meir},
  \bibinfo{author}{K.~Ciosek},
\newblock \bibinfo{title}{Discount factor as a regularizer in reinforcement
  learning},
\newblock in: \bibinfo{booktitle}{Proc. of the 37th International Conference on
  Machine Learning}, \bibinfo{year}{2020}.
\bibitem[{Ziebart et~al.(2008)Ziebart, Maas, Bagnell, and Dey}]{Ziebart2008a}
\bibinfo{author}{B.~D. Ziebart}, \bibinfo{author}{A.~Maas},
  \bibinfo{author}{J.~A. Bagnell}, \bibinfo{author}{A.~K. Dey},
\newblock \bibinfo{title}{Maximum entropy inverse reinforcement learning},
\newblock in: \bibinfo{booktitle}{Proc. of the 23rd AAAI Conference on
  Artificial Intelligence}, \bibinfo{year}{2008}.
\bibitem[{Belousov and Peters(2019)}]{Belousov2019a}
\bibinfo{author}{B.~Belousov}, \bibinfo{author}{J.~Peters},
\newblock \bibinfo{title}{Entropic regularization of {Markov} decision
  processes},
\newblock \bibinfo{journal}{Entropy} \bibinfo{volume}{21}
  (\bibinfo{year}{2019}) \bibinfo{pages}{3207--3245}.
\bibitem[{Ahmed et~al.(2019)Ahmed, N.~Le~Roux, and Schuurmans}]{Ahmed2019a}
\bibinfo{author}{Z.~Ahmed}, \bibinfo{author}{M.~N. N.~Le~Roux},
  \bibinfo{author}{D.~Schuurmans},
\newblock \bibinfo{title}{Understanding the impact of entropy on policy
  optimization},
\newblock in: \bibinfo{booktitle}{Proc. of the 36th International Conference on
  Machine Learning}, \bibinfo{year}{2019}, pp. \bibinfo{pages}{151--160}.
\bibitem[{Pomerleau(1989)}]{Pomerleau1989a}
\bibinfo{author}{D.~A. Pomerleau},
\newblock \bibinfo{title}{{ALVINN}: An autonomous land vehicle in a neural
  network},
\newblock in: \bibinfo{booktitle}{Advances in Neural Information Processing
  Systems 1}, \bibinfo{year}{1989}, pp. \bibinfo{pages}{305--313}.
\bibitem[{Laskey et~al.(2017)Laskey, Lee, Fox, Dragan, and
  Goldberg}]{Laskey2017a}
\bibinfo{author}{M.~Laskey}, \bibinfo{author}{J.~Lee},
  \bibinfo{author}{R.~Fox}, \bibinfo{author}{A.~Dragan},
  \bibinfo{author}{K.~Goldberg},
\newblock \bibinfo{title}{Dart: Noise injection for robust imitation learning},
\newblock in: \bibinfo{booktitle}{Proc. of the 1st Conference on Robot
  Learning}, \bibinfo{year}{2017}.
\bibitem[{Torabi et~al.(2018)Torabi, Warnell, and Stone}]{Torabi2018a}
\bibinfo{author}{F.~Torabi}, \bibinfo{author}{G.~Warnell},
  \bibinfo{author}{P.~Stone},
\newblock \bibinfo{title}{Behavioral cloning from observation},
\newblock in: \bibinfo{booktitle}{Proc. of the 27th International Joint
  Conference on Artificial Intelligence and the 23rd European Conference on
  Artificial Intelligence}, \bibinfo{year}{2018}, pp.
  \bibinfo{pages}{4950--57}.
\bibitem[{Reddy et~al.(2020)Reddy, Dragan, and Levine}]{Reddy2020a}
\bibinfo{author}{S.~Reddy}, \bibinfo{author}{A.~D. Dragan},
  \bibinfo{author}{S.~Levine},
\newblock \bibinfo{title}{Sqil: Imitation learning via regularized behavioral
  cloning},
\newblock in: \bibinfo{booktitle}{Proc. of the 8th International Conference on
  Learning Representations}, \bibinfo{year}{2020}.
\bibitem[{Kobayashi et~al.(2019)Kobayashi, Horii, Iwaki, Nagai, and
  Asada}]{Kobayashi2019a}
\bibinfo{author}{K.~Kobayashi}, \bibinfo{author}{T.~Horii},
  \bibinfo{author}{R.~Iwaki}, \bibinfo{author}{Y.~Nagai},
  \bibinfo{author}{M.~Asada},
\newblock \bibinfo{title}{Situated {GAIL}: Multitask imitation using
  task-conditioned adversarial inverse reinforcement learning},
\newblock \bibinfo{journal}{arXiv preprint}  (\bibinfo{year}{2019}).
\bibitem[{Henderson et~al.(2018)Henderson, Chang, Bacon, Meger, Pineau, and
  Precup}]{Henderson2018a}
\bibinfo{author}{P.~Henderson}, \bibinfo{author}{W.-D. Chang},
  \bibinfo{author}{P.-L. Bacon}, \bibinfo{author}{D.~Meger},
  \bibinfo{author}{J.~Pineau}, \bibinfo{author}{D.~Precup},
\newblock \bibinfo{title}{{OptionGAN}: Learning joint reward-policy options
  using generative adversarial inverse reinforcement learning},
\newblock in: \bibinfo{booktitle}{Proc. of the 32nd AAAI Conference on
  Artificial Intelligence}, \bibinfo{year}{2018}.
\bibitem[{Torabi et~al.(2019)Torabi, Warnell, and Stone}]{Torabi2019a}
\bibinfo{author}{F.~Torabi}, \bibinfo{author}{G.~Warnell},
  \bibinfo{author}{P.~Stone},
\newblock \bibinfo{title}{Generative adversarial imitation from observation},
\newblock in: \bibinfo{booktitle}{ICML 2019 Workshop on Imitation, Intent, and
  Interaction}, \bibinfo{year}{2019}.
\bibitem[{Sun and Ma(2014)}]{Sun2019a}
\bibinfo{author}{M.~Sun}, \bibinfo{author}{X.~Ma},
\newblock \bibinfo{title}{Adversarial imitation learning from incomplete
  demonstrations},
\newblock in: \bibinfo{booktitle}{Proc. of the 28th International Joint
  Conference on Artificial Intelligence}, \bibinfo{year}{2014}.
\bibitem[{Blond\'e and Kalousis(2019)}]{Blonde2019a}
\bibinfo{author}{L.~Blond\'e}, \bibinfo{author}{A.~Kalousis},
\newblock \bibinfo{title}{Sample-efficient imitation learning via generative
  adversarial nets},
\newblock in: \bibinfo{booktitle}{Proc. of the 22nd International Conference on
  Artificial Intelligence and Statistics}, \bibinfo{year}{2019}, pp.
  \bibinfo{pages}{3138--3148}.
\bibitem[{Lillicrap et~al.(2016)Lillicrap, Hunt, Pritzel, Heess, Erez, Tassa,
  Silver, and Wierstra}]{Lillicrap2016a}
\bibinfo{author}{T.~P. Lillicrap}, \bibinfo{author}{J.~J. Hunt},
  \bibinfo{author}{A.~Pritzel}, \bibinfo{author}{N.~Heess},
  \bibinfo{author}{T.~Erez}, \bibinfo{author}{Y.~Tassa},
  \bibinfo{author}{D.~Silver}, \bibinfo{author}{D.~Wierstra},
\newblock \bibinfo{title}{Continuous control with deep reinforcement learning},
\newblock in: \bibinfo{booktitle}{Proc. of the 4th International Conference on
  Learning Representations}, \bibinfo{address}{San Diego},
  \bibinfo{year}{2016}.
\bibitem[{Kostrikov et~al.(2019)Kostrikov, Agrawal, Dwibedi, Levine, and
  Tompson}]{Kostrikov2019a}
\bibinfo{author}{I.~Kostrikov}, \bibinfo{author}{K.~K. Agrawal},
  \bibinfo{author}{D.~Dwibedi}, \bibinfo{author}{S.~Levine},
  \bibinfo{author}{J.~Tompson},
\newblock \bibinfo{title}{Discriminator-actor-critic: Addressing sample
  inefficiency and reward bias in adversarial imitation learning},
\newblock in: \bibinfo{booktitle}{Proc. of the 7th International Conference on
  Learning Representations}, \bibinfo{year}{2019}.
\bibitem[{Fujimoto et~al.(2018)Fujimoto, van Hoof, and Meger}]{Fujimoto2018a}
\bibinfo{author}{S.~Fujimoto}, \bibinfo{author}{H.~van Hoof},
  \bibinfo{author}{D.~Meger},
\newblock \bibinfo{title}{Addressing function approximation error in
  actor-critic methods},
\newblock in: \bibinfo{booktitle}{Proc. of the 35th International Conference on
  Machine Learning}, \bibinfo{year}{2018}.
\bibitem[{Sasaki et~al.(2019)Sasaki, Yohira, and Kawaguchi}]{Sasaki2019a}
\bibinfo{author}{F.~Sasaki}, \bibinfo{author}{T.~Yohira},
  \bibinfo{author}{A.~Kawaguchi},
\newblock \bibinfo{title}{Sample efficient imitation learning for continuous
  control},
\newblock in: \bibinfo{booktitle}{Proc. of the 7th International Conference on
  Learning Representations}, \bibinfo{year}{2019}.
\bibitem[{Degris et~al.(2012)Degris, White, and Sutton}]{Degris2012a}
\bibinfo{author}{T.~Degris}, \bibinfo{author}{M.~White}, \bibinfo{author}{R.~S.
  Sutton},
\newblock \bibinfo{title}{Off-policy actor-critic},
\newblock in: \bibinfo{booktitle}{Proc. of the 29th International Conference on
  Machine Learning}, \bibinfo{year}{2012}.
\bibitem[{Zuo et~al.(2020)Zuo, Chen, Lu, and Huang}]{Zuo2020a}
\bibinfo{author}{G.~Zuo}, \bibinfo{author}{K.~Chen}, \bibinfo{author}{J.~Lu},
  \bibinfo{author}{X.~Huang},
\newblock \bibinfo{title}{Deterministic generative adversarial imitation
  learning},
\newblock \bibinfo{journal}{Neurocomputing}  (\bibinfo{year}{2020})
  \bibinfo{pages}{60--69}.
\bibitem[{Nishio et~al.(2020)Nishio, Kuyoshi, Tsuneda, and
  Yamane}]{Nishio2020a}
\bibinfo{author}{D.~Nishio}, \bibinfo{author}{D.~Kuyoshi},
  \bibinfo{author}{T.~Tsuneda}, \bibinfo{author}{S.~Yamane},
\newblock \bibinfo{title}{Discriminator soft actor critic without extrinsic
  rewards},
\newblock \bibinfo{journal}{arXiv preprint}  (\bibinfo{year}{2020}).
\bibitem[{Ghasemipour et~al.(2019)Ghasemipour, Zemel, and
  Gu}]{Ghasemipour2019a}
\bibinfo{author}{S.~K.~S. Ghasemipour}, \bibinfo{author}{R.~Zemel},
  \bibinfo{author}{S.~Gu},
\newblock \bibinfo{title}{A divergence minimization perspective on imitation
  learning methods},
\newblock in: \bibinfo{booktitle}{Proc. of the 3rd Conference on Robot
  Learning}, \bibinfo{year}{2019}, pp. \bibinfo{pages}{1259--1277}.
\bibitem[{Ke et~al.(2020)Ke, Barnes, Sun, Lee, Choudhury, and
  Srinivasa}]{Ke2020a}
\bibinfo{author}{L.~Ke}, \bibinfo{author}{M.~Barnes}, \bibinfo{author}{W.~Sun},
  \bibinfo{author}{G.~Lee}, \bibinfo{author}{S.~Choudhury},
  \bibinfo{author}{S.~Srinivasa},
\newblock \bibinfo{title}{Imitation learning as $f$-divergence minimization},
\newblock in: \bibinfo{booktitle}{Proc. of the 14th International Workshop on
  the Algorithmic Foundations of Robotics (WAFR)}, \bibinfo{year}{2020}.
\bibitem[{Peters and Schaal(2008)}]{Peters2008a}
\bibinfo{author}{J.~Peters}, \bibinfo{author}{S.~Schaal},
\newblock \bibinfo{title}{Reinforcement learning of motor skills with policy
  gradients},
\newblock \bibinfo{journal}{Neural Networks}  (\bibinfo{year}{2008})
  \bibinfo{pages}{1--13}.
\bibitem[{Dieng et~al.(2019)Dieng, Ruiz, Blei, and Titsias}]{Dieng2019a}
\bibinfo{author}{A.~B. Dieng}, \bibinfo{author}{F.~J.~R. Ruiz},
  \bibinfo{author}{D.~M. Blei}, \bibinfo{author}{M.~K. Titsias},
\newblock \bibinfo{title}{Prescribed generative adversarial networks},
\newblock \bibinfo{journal}{arXiv preprint}  (\bibinfo{year}{2019}).
\bibitem[{Schaul et~al.(2015)Schaul, Horgan, Gregor, and Silver}]{Schaul2015a}
\bibinfo{author}{T.~Schaul}, \bibinfo{author}{D.~Horgan},
  \bibinfo{author}{K.~Gregor}, \bibinfo{author}{D.~Silver},
\newblock \bibinfo{title}{Universal value function approximators},
\newblock in: \bibinfo{booktitle}{Proc. of the 32nd International Conference on
  Machine Learning}, \bibinfo{year}{2015}, pp. \bibinfo{pages}{1312--1320}.
\bibitem[{Kingma and Ba(2015)}]{Kingma2015a}
\bibinfo{author}{D.~Kingma}, \bibinfo{author}{J.~Ba},
\newblock \bibinfo{title}{{ADAM}: A method for stochastic optimization},
\newblock in: \bibinfo{booktitle}{Proc. of the 3rd International Conference for
  Learning Representations}, \bibinfo{year}{2015}.
\bibitem[{Chitta et~al.(2012)Chitta, Sucan, and Cousins}]{Chitta2012a}
\bibinfo{author}{S.~Chitta}, \bibinfo{author}{I.~Sucan},
  \bibinfo{author}{S.~Cousins},
\newblock \bibinfo{title}{Moveit! [{ROS} topics]},
\newblock \bibinfo{journal}{IEEE Robotics Automation Magazine}
  \bibinfo{volume}{19} (\bibinfo{year}{2012}) \bibinfo{pages}{18--19}.
\bibitem[{Yarats et~al.(2020)Yarats, Zhang, Kostrikov, Amos, Pineau, and
  Fergus}]{Yarats2020a}
\bibinfo{author}{D.~Yarats}, \bibinfo{author}{A.~Zhang},
  \bibinfo{author}{I.~Kostrikov}, \bibinfo{author}{B.~Amos},
  \bibinfo{author}{J.~Pineau}, \bibinfo{author}{R.~Fergus},
\newblock \bibinfo{title}{Improving sample efficiency in model-free
  reinforcement learning from images},
\newblock \bibinfo{journal}{arXiv preprint}  (\bibinfo{year}{2020}).
\bibitem[{Ghosh et~al.(2020)Ghosh, Sajjadi, Vergari, Black, and
  Scholkopf}]{Ghosh2020a}
\bibinfo{author}{P.~Ghosh}, \bibinfo{author}{M.~S.~M. Sajjadi},
  \bibinfo{author}{A.~Vergari}, \bibinfo{author}{M.~Black},
  \bibinfo{author}{B.~Scholkopf},
\newblock \bibinfo{title}{From variational to deterministic autoencoders},
\newblock in: \bibinfo{booktitle}{Proc. of the 7th International Conference on
  Learning Representations}, \bibinfo{year}{2020}.
\bibitem[{Wang(2012)}]{Wang2012a}
\bibinfo{author}{J.-J. Wang},
\newblock \bibinfo{title}{Stabilization and tracking control of x-z inverted
  pendulum with sliding-mode control},
\newblock \bibinfo{journal}{ISA Transactions} \bibinfo{volume}{51}
  (\bibinfo{year}{2012}) \bibinfo{pages}{763--70}.
\bibitem[{Henderson et~al.(2018)Henderson, Islam, Bachman, Pineau, Precup, and
  Meger}]{Henderson2018b}
\bibinfo{author}{P.~Henderson}, \bibinfo{author}{R.~Islam},
  \bibinfo{author}{P.~Bachman}, \bibinfo{author}{J.~Pineau},
  \bibinfo{author}{D.~Precup}, \bibinfo{author}{D.~Meger},
\newblock \bibinfo{title}{Deep reinforcement learning that matters},
\newblock in: \bibinfo{booktitle}{Proc. of the 32nd AAAI Conference on
  Artificial Intelligence}, \bibinfo{year}{2018}.
\bibitem[{Zhang et~al.(2021)Zhang, Rajan, Pineda, Lambert, Biedenkapp, Chua,
  Hutter, and Calandra}]{Zhang2021a}
\bibinfo{author}{B.~Zhang}, \bibinfo{author}{R.~Rajan},
  \bibinfo{author}{L.~Pineda}, \bibinfo{author}{N.~Lambert},
  \bibinfo{author}{A.~Biedenkapp}, \bibinfo{author}{K.~Chua},
  \bibinfo{author}{F.~Hutter}, \bibinfo{author}{R.~Calandra},
\newblock \bibinfo{title}{On the importance of hyperparameter optimization for
  model-based reinforcement learning},
\newblock in: \bibinfo{booktitle}{Proc. of the 24th International Conference on
  Artificial Intelligence and Statistics}, \bibinfo{year}{2021}, pp.
  \bibinfo{pages}{4015--23}.
\bibitem[{Elfwing et~al.(2018)Elfwing, E.Uchibe, and Doya}]{Elfwing2018b}
\bibinfo{author}{S.~Elfwing}, \bibinfo{author}{E.Uchibe},
  \bibinfo{author}{K.~Doya},
\newblock \bibinfo{title}{Online meta-learning by parallel algorithm
  competition},
\newblock in: \bibinfo{booktitle}{Proc. of the Genetic and Evolutionary
  Computation Conference}, \bibinfo{year}{2018}, pp. \bibinfo{pages}{426--33}.
\bibitem[{Jaderberg et~al.(2017)Jaderberg, Dalibard, Osindero, Czarnecki,
  Donahue, Razavi, Vinyals et~al.}]{Jaderberg2017a}
\bibinfo{author}{M.~Jaderberg}, \bibinfo{author}{V.~Dalibard},
  \bibinfo{author}{S.~Osindero}, \bibinfo{author}{W.~M. Czarnecki},
  \bibinfo{author}{J.~Donahue}, \bibinfo{author}{A.~Razavi},
  \bibinfo{author}{O.~Vinyals}, et~al.,
\newblock \bibinfo{title}{Population based training of neural networks},
\newblock \bibinfo{journal}{arXiv preprint}  (\bibinfo{year}{2017}).
\bibitem[{Lee et~al.(2020)Lee, Lee, Vrancx, Kim, and Kim}]{Lee2020a}
\bibinfo{author}{B.-J. Lee}, \bibinfo{author}{J.~Lee},
  \bibinfo{author}{P.~Vrancx}, \bibinfo{author}{D.~Kim}, \bibinfo{author}{K.-E.
  Kim},
\newblock \bibinfo{title}{Batch reinforcement learning with hyperparameter
  gradients},
\newblock in: \bibinfo{booktitle}{Proc. of the 37th International Conference on
  Machine Learning}, \bibinfo{year}{2020}.

\end{thebibliography}

\clearpage
\renewcommand{\thefigure}{\arabic{figure}}

\begin{figure}[t]
  \centering
  \includegraphics[width=1.0\linewidth]{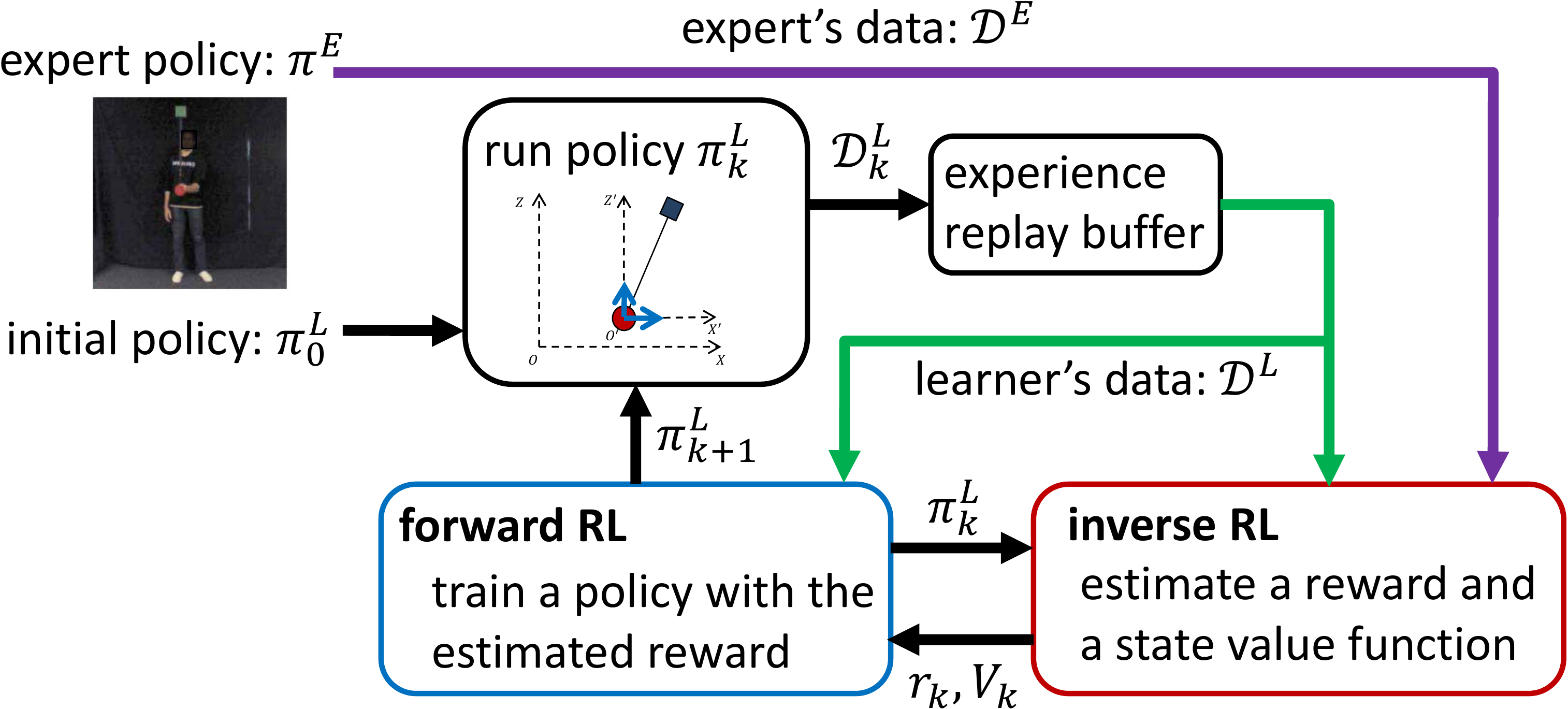}
  \caption{Overall architecture of Entropy-Regularized Imitation
    Learning (ERIL)}
    \label{fig:eril_interaction}
\end{figure}

\begin{figure}[t]
  \centering
  \includegraphics[width=1.0\linewidth]{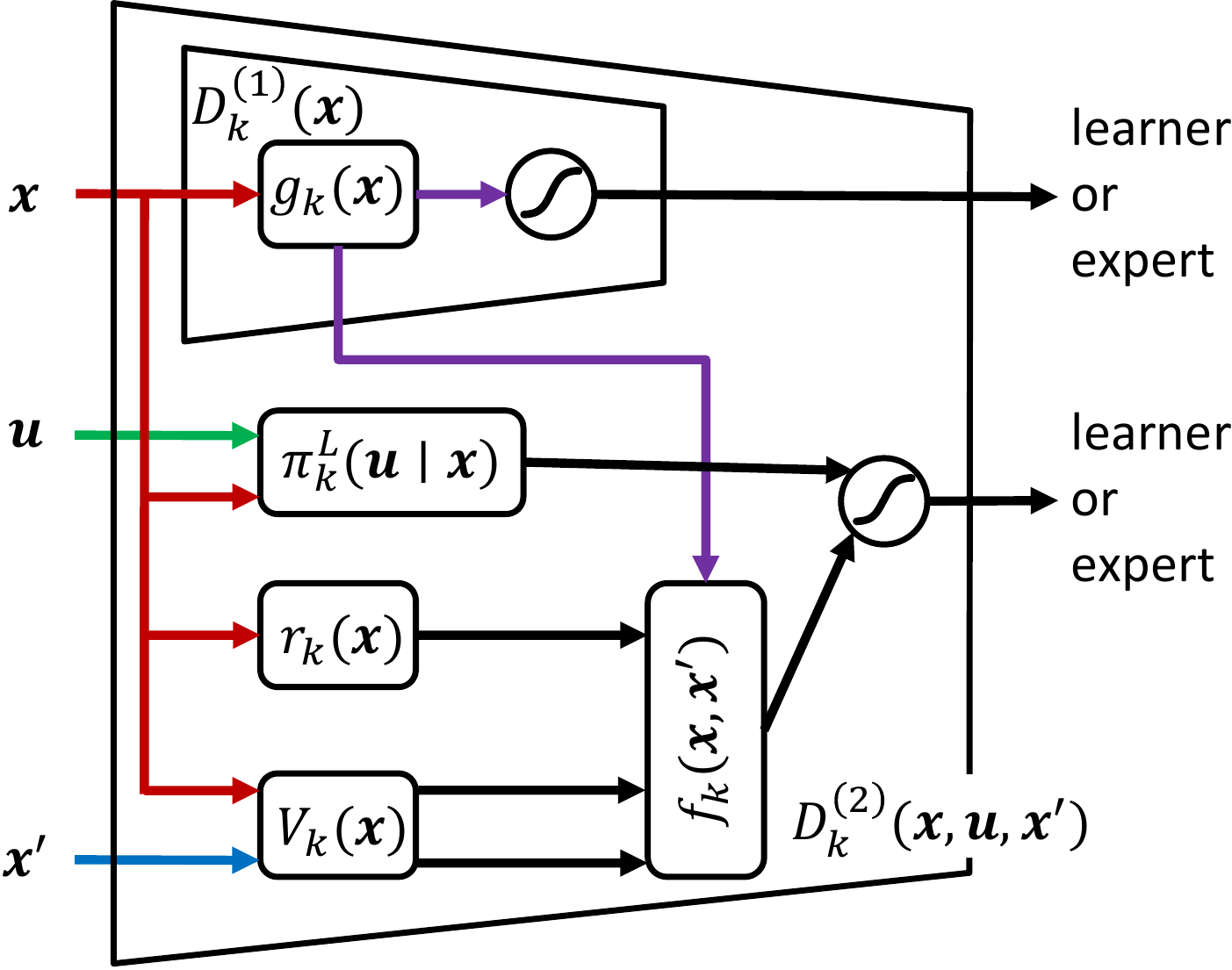}
  \caption{Architecture of two ERIL discriminators: $D_k^{(1)}(\bx)$
    and $D_k^{(2)}(\bx, \bu, \bx')$.}
    \label{fig:eril_discriminator}
\end{figure}

\begin{figure}[t]
  \centering
  \includegraphics[width=1.0\linewidth]{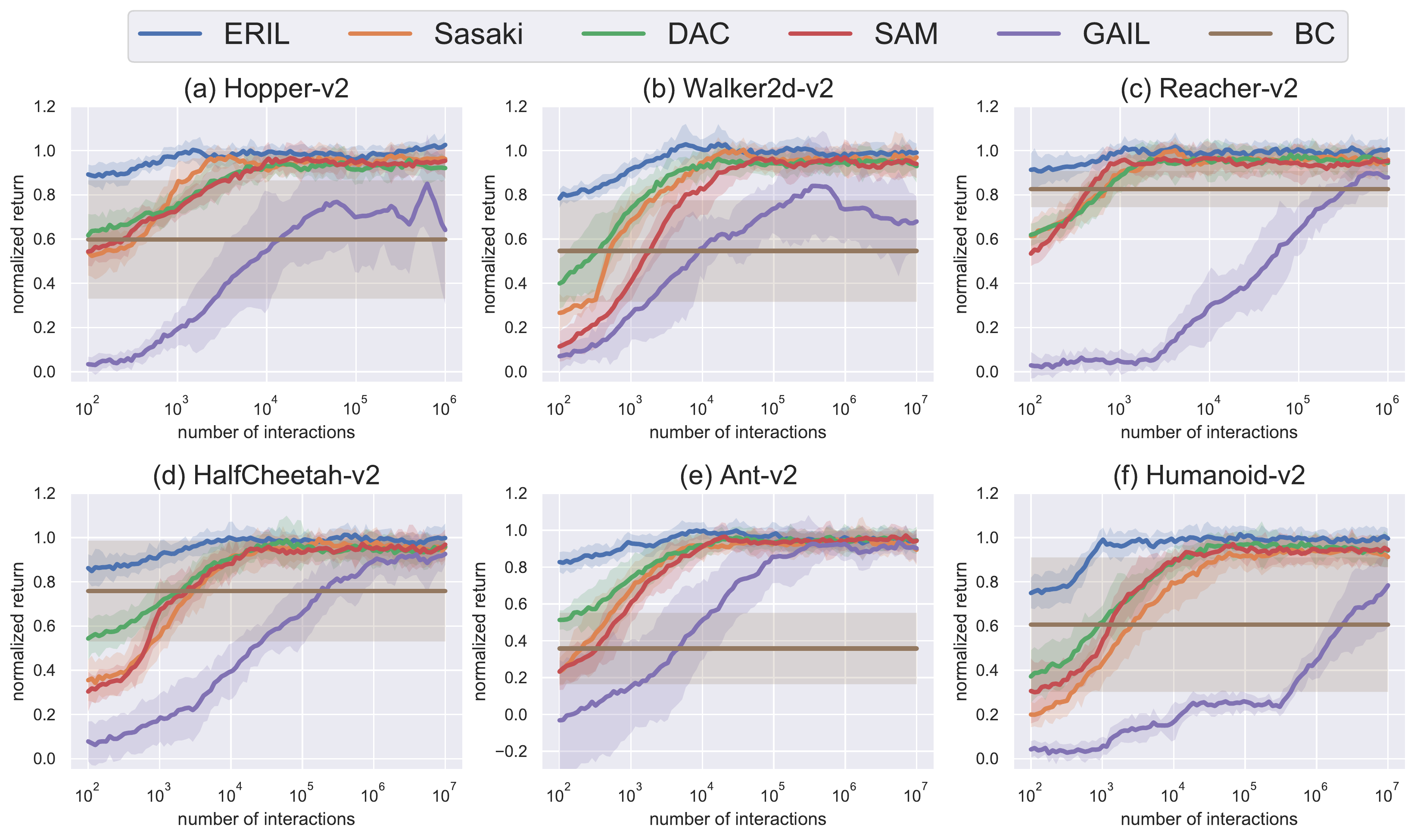}
  \caption{Performance with respect to number of interactions
    in MuJoCo tasks:
    Solid lines represent average values, and shaded areas
    correspond to $\pm 1$ standard deviation region.}
    \label{fig:mujoco:frl_normal}
\end{figure}

\begin{figure}[t]
  \centering
  \includegraphics[width=1.0\linewidth]{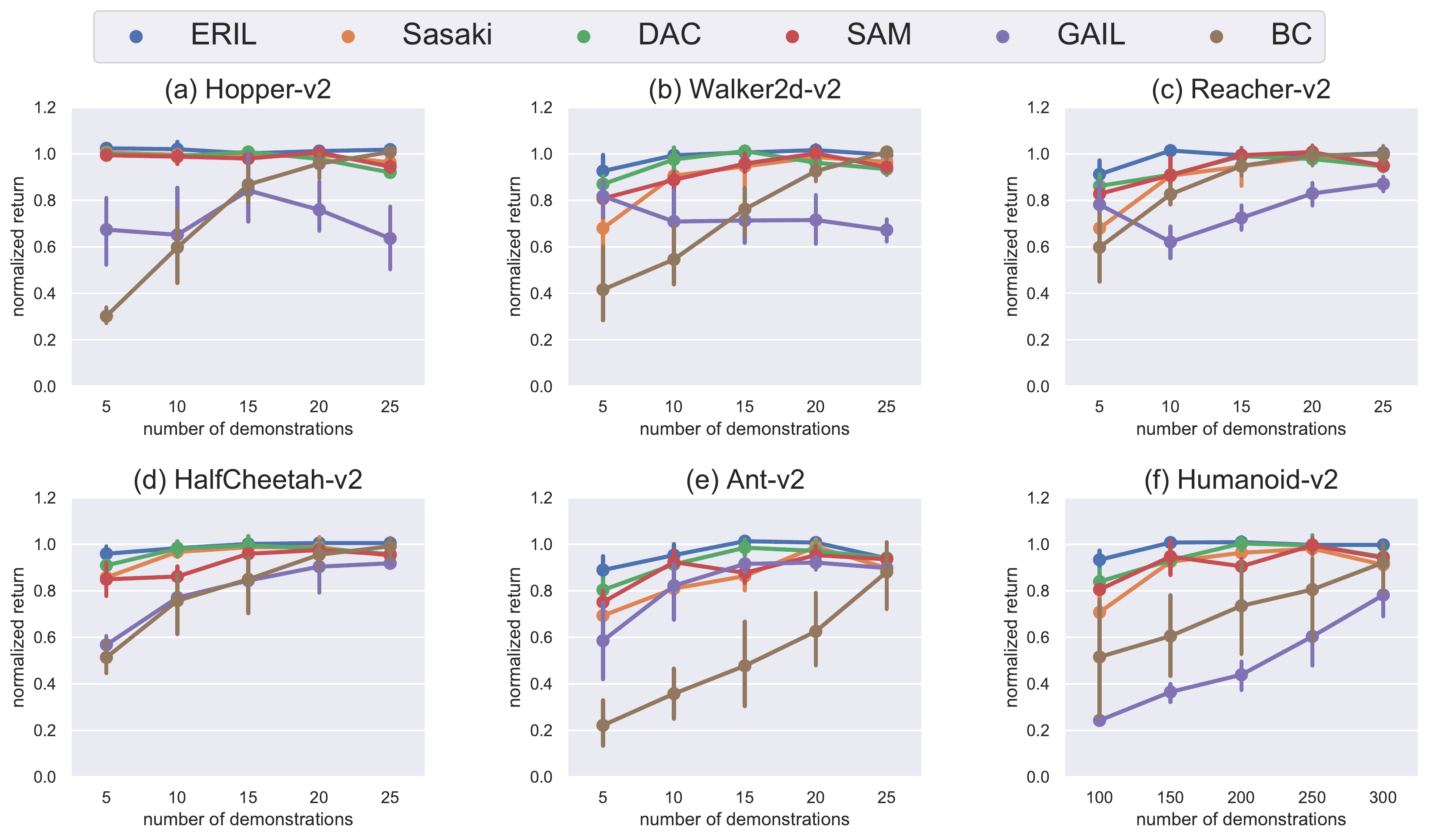} 
  \caption{Performance with respect to number of trajectories
    provided by expert in MuJoCo tasks:
    Solid lines represent average values, and shaded areas
    correspond to $\pm 1$ standard deviation region.}
  \label{fig:mujoco:irl_normal}
\end{figure}

\begin{figure}[t]
  \centering
  \includegraphics[width=1.0\linewidth]{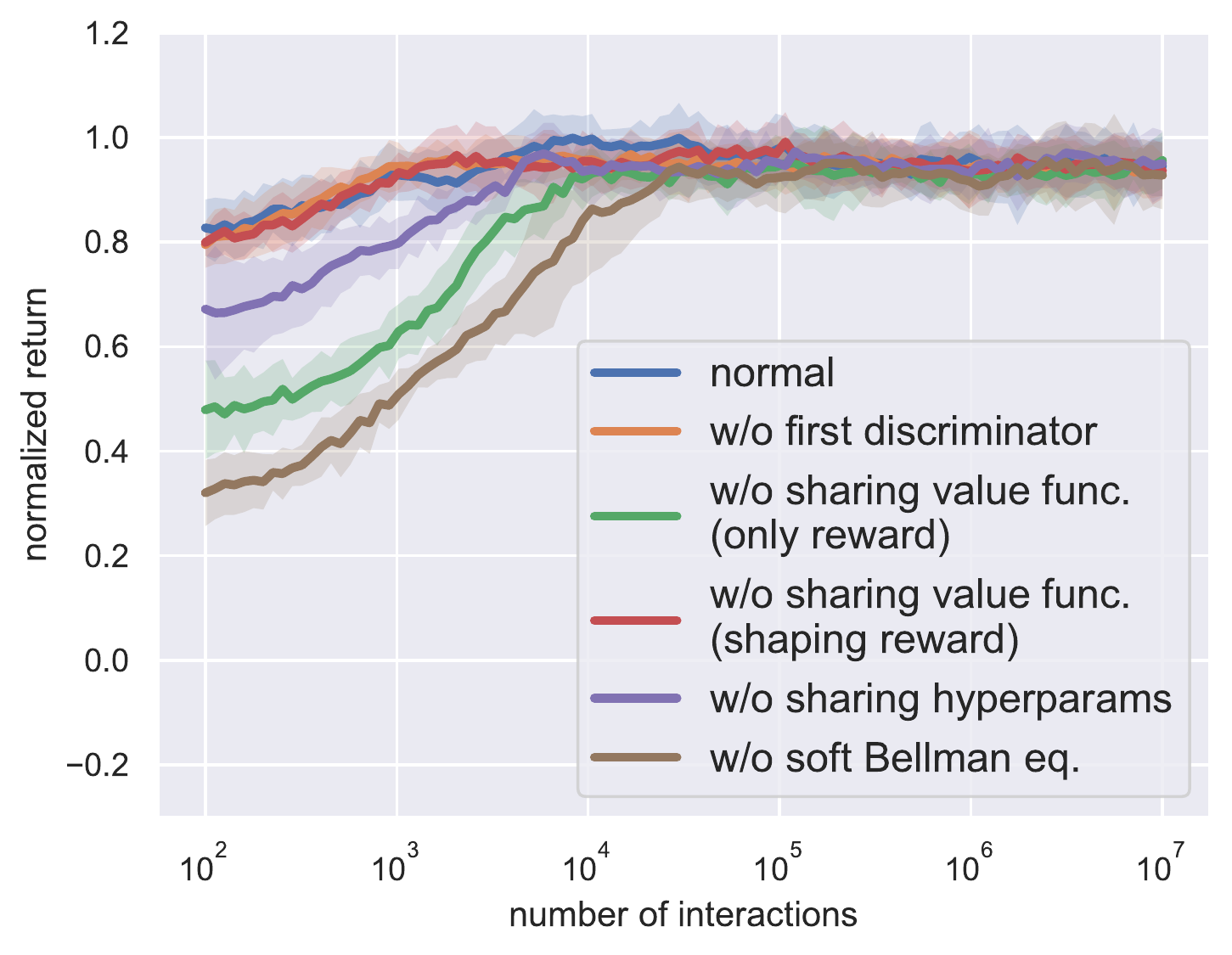}
  \caption{Comparison of learning curves in ablation study}
    \label{fig:mujoco:frl_ablation}
\end{figure}

\begin{figure}[t]
  \centering
  \includegraphics[width=1.0\hsize]{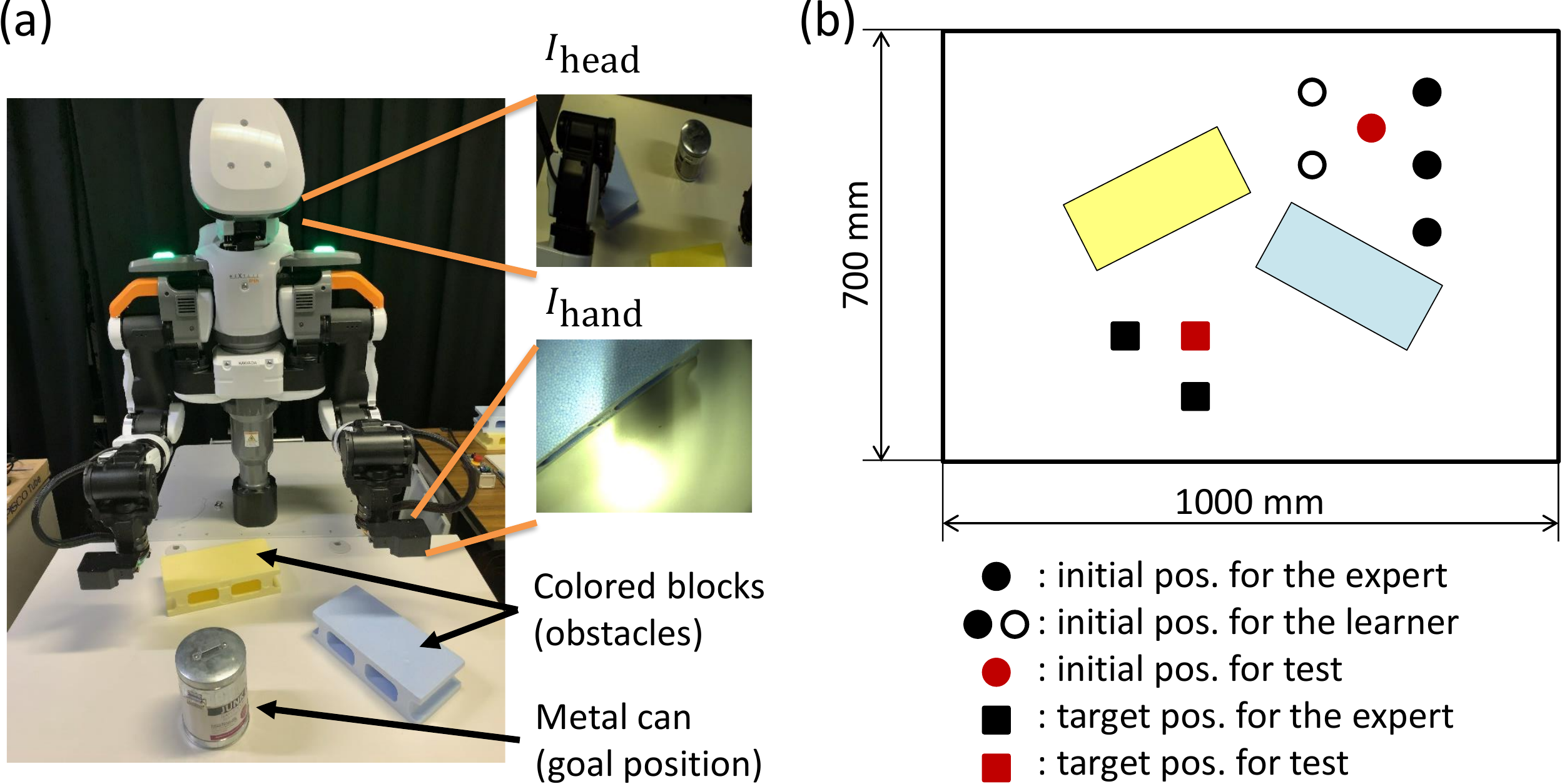}
  \caption{Real robot setting: (a) environment of reaching task, which
   moves end-effector of left arm to 
    target position; (b) possible environmental configuration.}
  \label{fig:nextage:task}
\end{figure}

\begin{figure}[t]
  \centering
  \includegraphics[width=1.0\hsize]{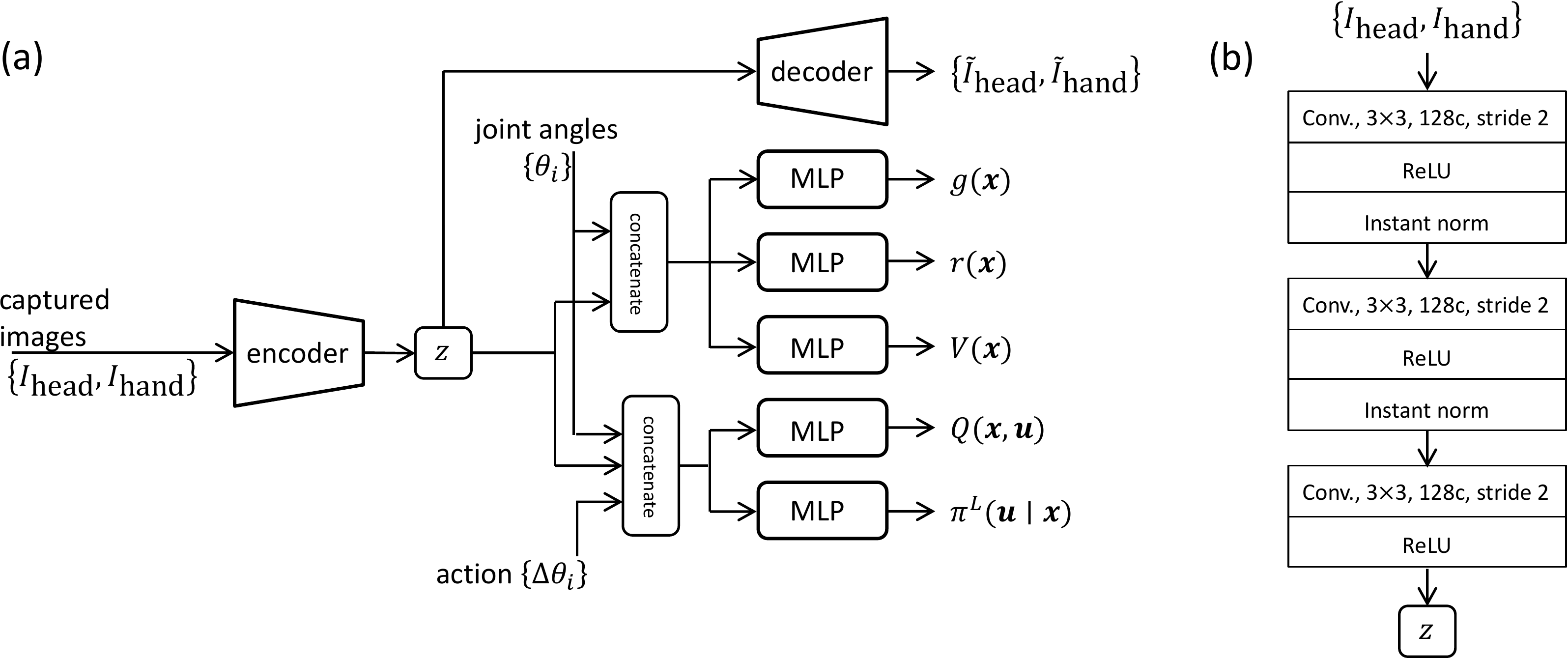}
  \caption{Neural networks used in robot experiment:
    (a) architecture for policy, reward, state value,
    state-action value, and first discriminator, sharing
    the encoder;
    (b) architecture for encoder: Conv. denotes a
    convolutional neural network. ``$n$c'' denotes
   ``$n$ channels.''}
  \label{fig:nextage:network}
\end{figure}

\begin{figure}[t]
  \centering
  \includegraphics[width=1.0\hsize]{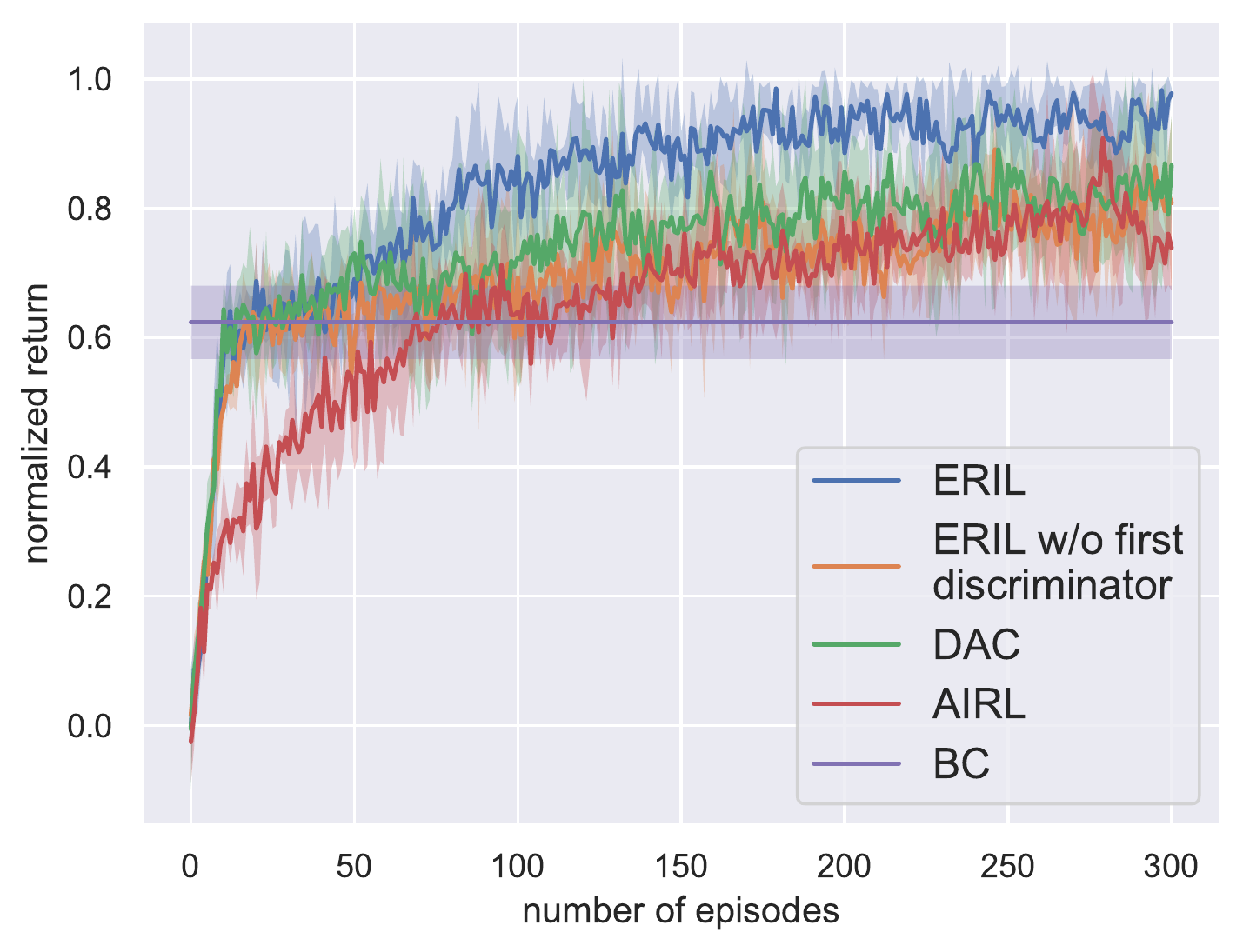}
  \caption{Performance with respect to number of interactions
    in real robot experiment}
  \label{fig:nextage:performance}
\end{figure}

\begin{figure}[t]
  \centering
  \includegraphics[width=1.0\hsize]{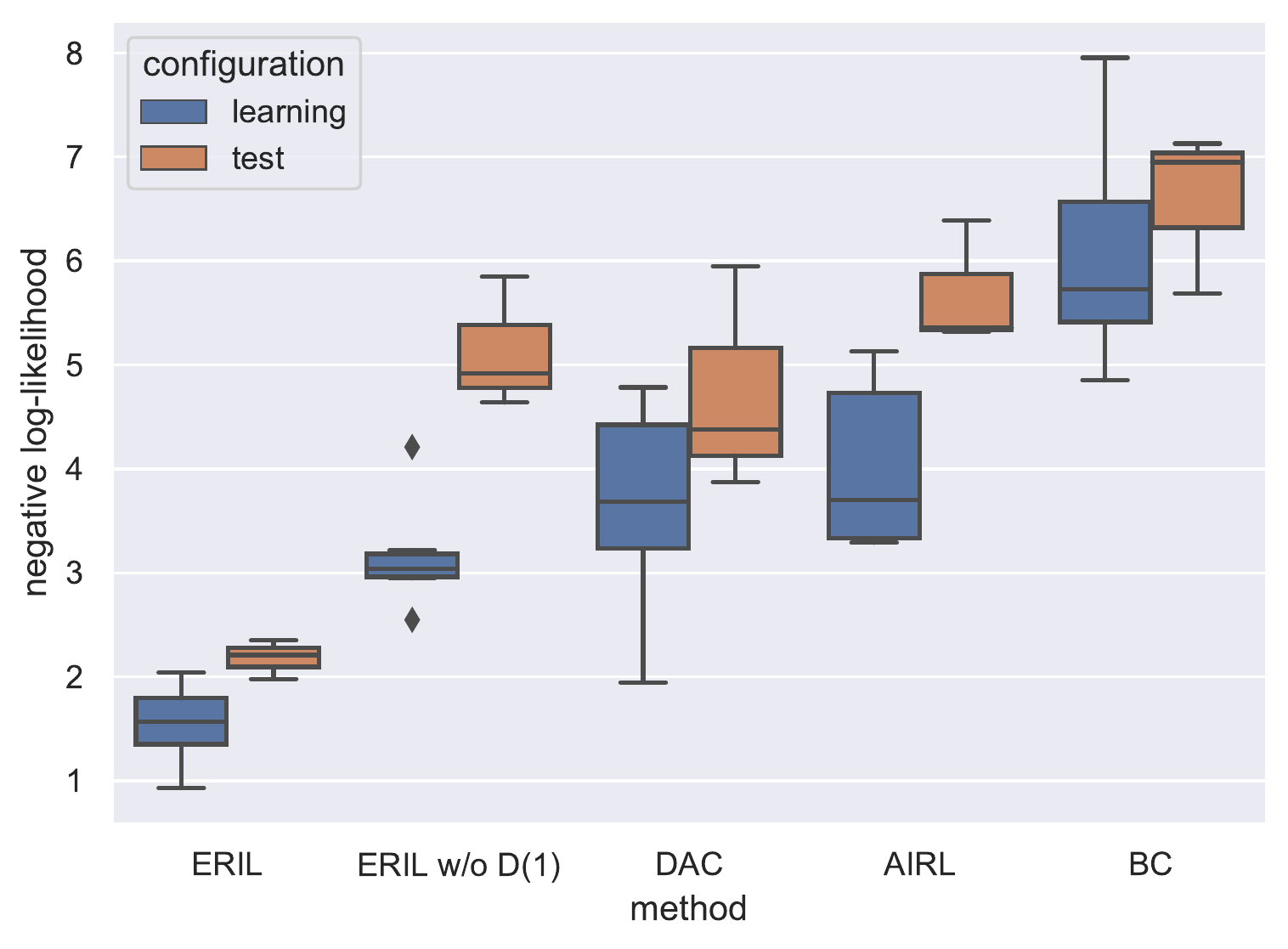}
  \caption{Comparison of NLL in real robot experiment:
    ``ERIL w/o D(1)'' denotes ERIL without first discriminator.
    Note that smaller NLL values indicate a better fit.}
  \label{fig:nextage:nll}
\end{figure}

\begin{figure}[t]
  \centering
  \includegraphics[width=1.0\hsize]{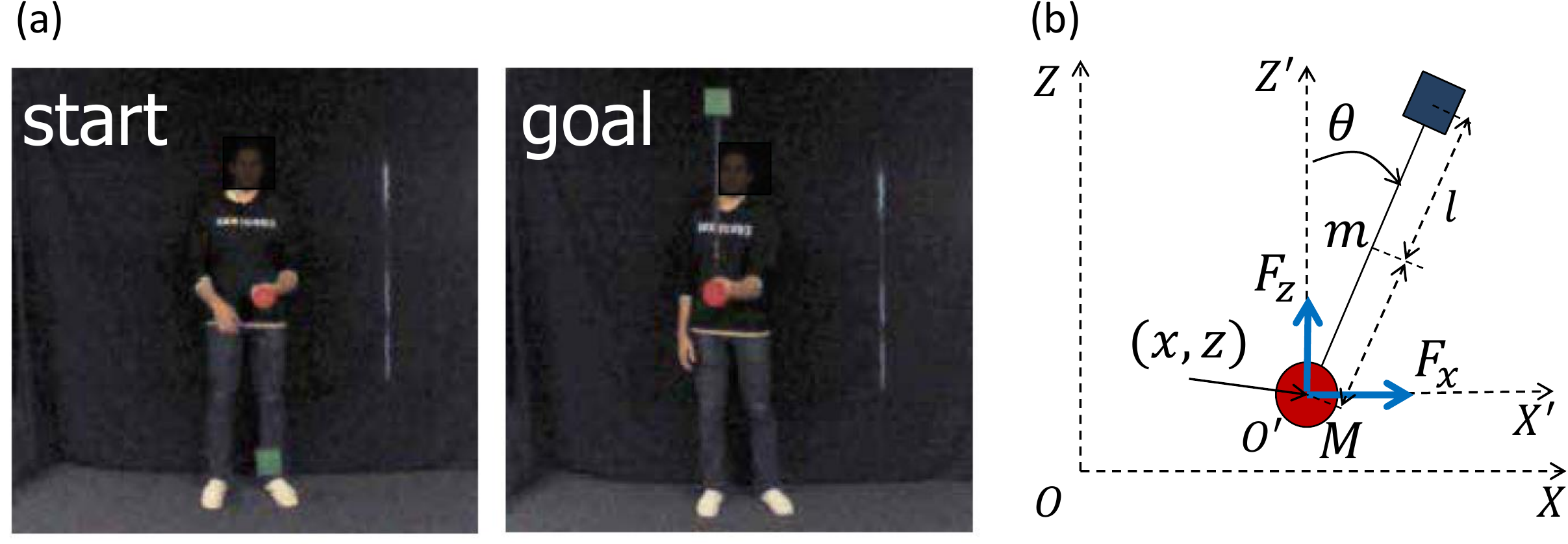}
  \caption{Inverted pendulum task solved by a human subject:
    (a) start and goal positions; (b) state representation.
    Notations are explained in
      \ref{sec:xz_inverted_pendulum}.}
  \label{fig:human:task}
\end{figure}

\begin{figure}[t]
  \centering
  \includegraphics[width=1.0\hsize]{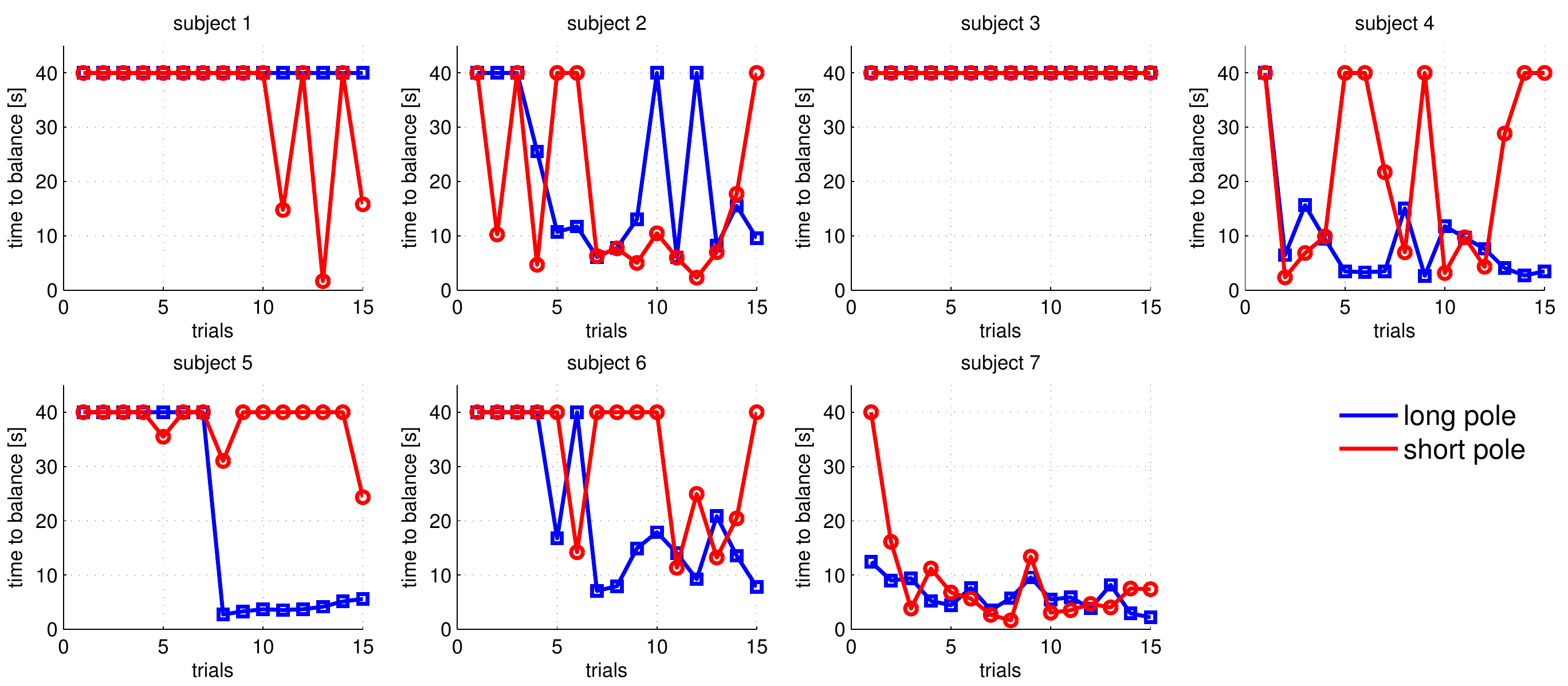}
  \caption{Learning curves of seven subjects: blue: long pole;
    red: short pole. Trial is considered a failure if subject
    cannot balance pole in upright position within 40 [s].}
  \label{fig:human:performance}
\end{figure}

\begin{figure}[t]
  \centering
  \includegraphics[width=1.0\hsize]{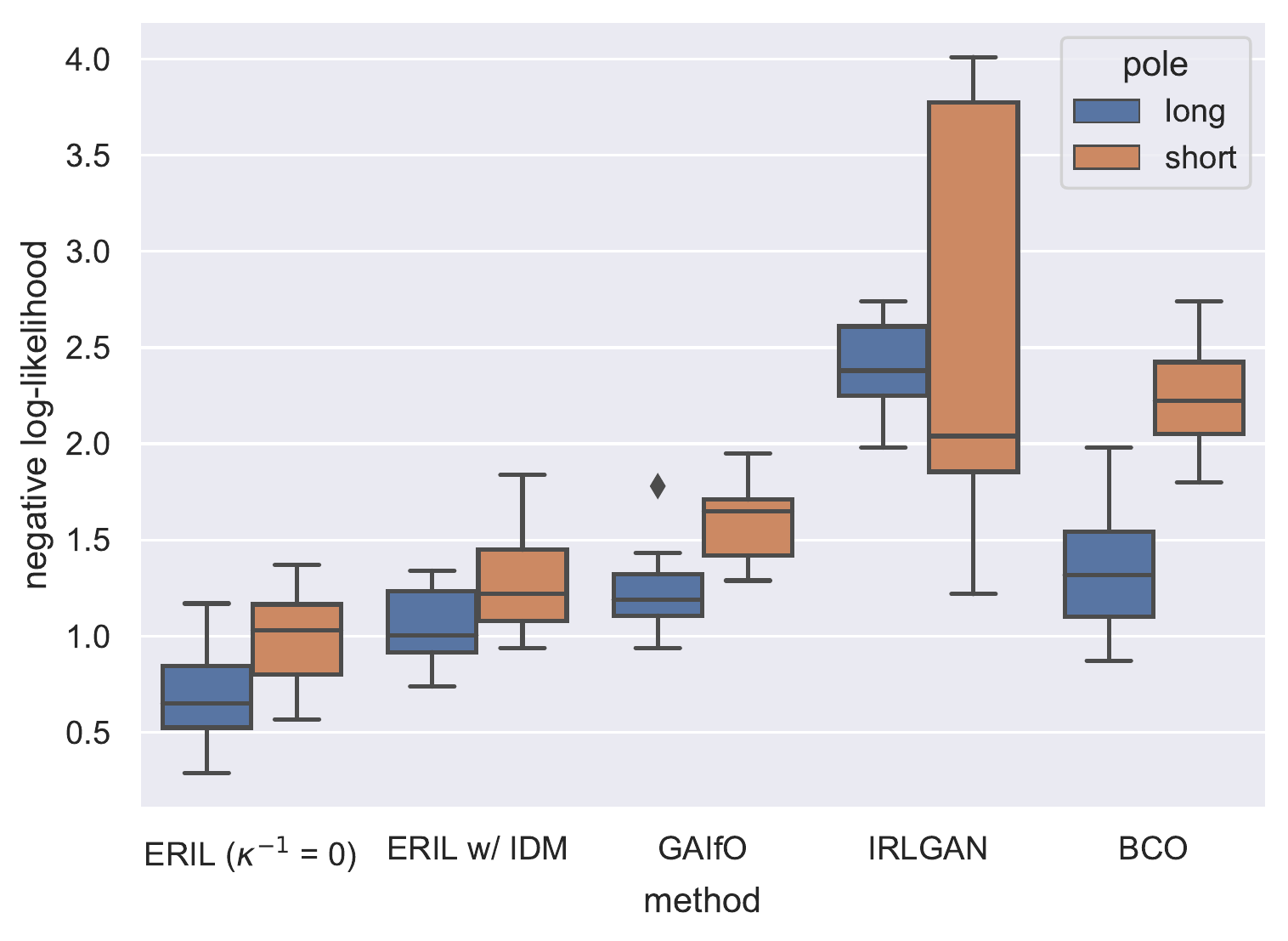}
  \caption{Comparison of NLL among imitation
    learning algorithms: Note that smaller NLL values 
    indicate a better fit.}
  \label{fig:human:nll_methods}
\end{figure}

\begin{figure}[t]
  \centering
  \includegraphics[width=1.0\hsize]{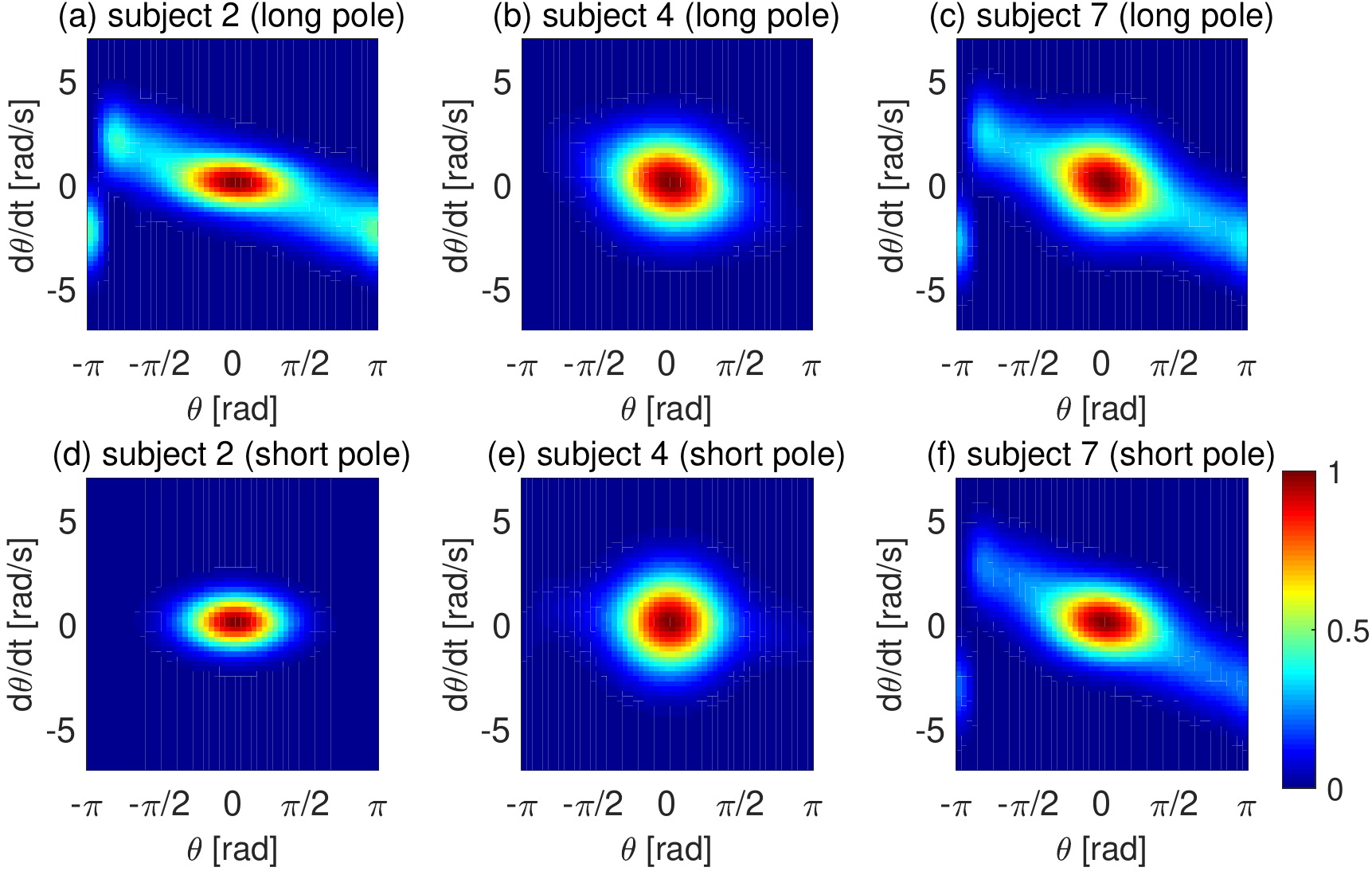}
  \caption{Estimated reward function of Subjects 2, 4, and 7
    projected to subspace
    $(\theta, \omega)$, $\omega = d\theta/dt$}
  \label{fig:human:rewards}
\end{figure}

\begin{figure}[t]
  \centering
  \includegraphics[width=1.0\hsize]{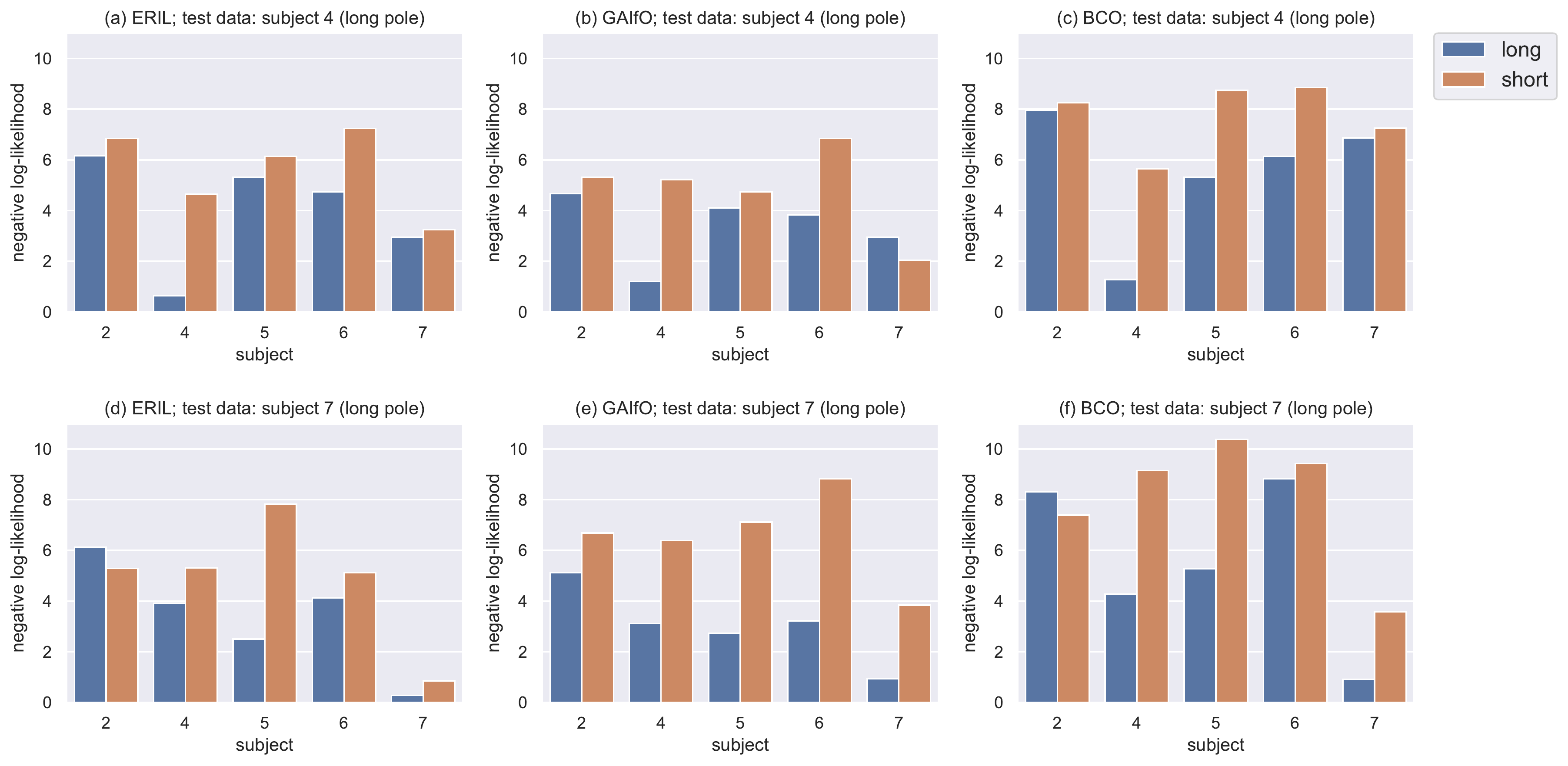}
  \caption{Comparison of NLL when condition of
    test dataset was different from training one 
    in human inverted pendulum task. 
    Figs. in upper row (a, b, and c) show results when 
    test dataset is $\mathcal{D}_{4, 1, \mathrm{te}}^E$.
    Figs. in lower row (d, e, and f) show results when 
    test dataset is $\mathcal{D}_{7, 1, \mathrm{te}}^E$.}
  \label{fig:human:nll_subjects}
\end{figure}

\end{document}